\documentclass{article}

\PassOptionsToPackage{numbers,compress}{natbib}
\usepackage[preprint]{neurips_2026}

\usepackage[utf8]{inputenc}
\usepackage[T1]{fontenc}
\usepackage{hyperref}
\usepackage{url}
\usepackage{graphicx}
\usepackage{amsmath}
\usepackage{booktabs}
\usepackage{multirow}
\usepackage{amsfonts}
\usepackage{nicefrac}
\usepackage{microtype}
\usepackage{xcolor}
\usepackage{capt-of}
\usepackage{algorithm}
\usepackage{algpseudocode}
\usepackage{float}

\title{EA-WM: Event-Aware Generative World Model with Structured Kinematic-to-Visual Action Fields}

\author{%
\textbf{Zhaoyang Yang}$^{\textbf{1,3}}$ \qquad
\textbf{Yurun Jin}$^{\textbf{3,4}}$ \\[0.2em]
\textbf{Lizhe Qi}$^{\textbf{1}^{\textbf{*}}}$ \qquad
\textbf{Cong Huang}$^{\textbf{2,3}^{*}}$ \qquad
\textbf{Kai Chen}$^{\textbf{2,3,5}^{*\dagger}}$ \\[0.6em]
$^{1}$Fudan University \qquad
$^{2}$Zhongguancun Academy \\
$^{3}$Zhongguancun Institute of Artificial Intelligence \\
$^{4}$University of Science and Technology of China \qquad
$^{5}$DeepCybo
}

\begin{document}

\maketitle

\begingroup
\renewcommand{\thefootnote}{}
\footnotetext{\textsuperscript{*}Corresponding authors. \textsuperscript{\textdagger}Project leader.}
\endgroup

\begin{center}
\small\url{https://github.com/Shownx-c/EA-WM}
\end{center}

\begin{abstract}
Pretrained video diffusion models provide powerful spatiotemporal generative priors, making them a natural foundation for robotic world models. While recent world-action models jointly optimize future videos and actions, they predominantly treat video generation as an auxiliary representation for policy learning. Consequently, they insufficiently explore the inverse problem: leveraging action signals to guide video synthesis, thereby often failing to preserve precise robot spatial geometry and fine-grained robot-object interaction dynamics in the generated rollouts. To bridge this gap, we present \textbf{EA-WM}, an Event-Aware Generative World Model that effectively closes the loop between kinematic control and visual perception. Rather than injecting joint or end-effector actions as abstract, low-dimensional tokens, EA-WM projects actions and kinematic states directly into the target camera view as \textit{Structured Kinematic-to-Visual Action Fields} (KVAFs). To fully exploit this geometrically grounded representation, we introduce event-aware bidirectional fusion blocks that modulate cross-branch attention, capturing object state changes and interaction dynamics. Evaluated on the comprehensive WorldArena benchmark, EA-WM achieves state-of-the-art performance, outperforming existing baselines by a significant margin.
\end{abstract}

\section{Introduction}
Recent video foundation models have rapidly improved the ability to synthesize temporally coherent, high-fidelity videos. From latent video diffusion models and space-time diffusion architectures to diffusion-transformer systems such as CogVideoX \citep{yang2025cogvideox} and the open Wan series \citep{wanteam2025wan,wan2025wan22,gao2025wans2v}, these models learn strong priors over appearance, motion, and scene evolution from large-scale video corpora \citep{blattmann2023stablevideodiffusion,bartal2024lumiere,yang2025cogvideox,wanteam2025wan}. This progress has motivated the view that scaled video generation can move beyond visual content creation toward world simulation \citep{openai2024sora,agarwal2025cosmos}. Since robot manipulation is fundamentally a process of predicting how actions change visual states over time, such spatiotemporal priors make video generation models a natural foundation for robotic world modeling.

Driven by this potential, recent robotics research has begun to adapt these video priors for several roles. Some methods \citep{wu2024ivideogpt,rigter2024avid,zhu2025irasim} use video world models as predictive simulators to roll out the visual consequences of candidate actions, enabling action-conditioned video prediction, visual planning, model-based reinforcement learning, and fine-grained robot-object interaction simulation. Other work \citep{jang2025dreamgen,liang2025videopolicy,quevedo2025worldgym,li2025worldeval,tseng2025scalablepolicyeval} uses generated rollouts as robot-learning infrastructure: they can synthesize additional robot trajectories beyond manual data collection, provide video-based supervision for policy learning, and offer scalable proxies for evaluating VLA policies without repeatedly executing every policy on physical robots.

Meanwhile, recent world-action models \citep{zhu2025uwm,ye2026dreamzero,kim2026cosmospolicy} have begun to jointly tune video generation and action modeling within unified generative frameworks, showing that future videos can provide dense world representations for action generation, policy learning, planning, and value estimation.  However, these works predominantly treat video generation merely as an auxiliary representation to optimize action prediction. Consequently, they insufficiently explore the crucial inverse problem: how to effectively leverage control signals to guide accurate video synthesis. This oversight leads to a critical limitation: the generated videos often fail to preserve the precise spatial geometry of robot motion and the complex dynamics of robot-object interactions. As high-quality video rollouts are increasingly relied upon as predictive simulators and synthetic data engines for downstream tasks \citep{jang2025dreamgen,liang2025videopolicy,quevedo2025worldgym,li2025worldeval,tseng2025scalablepolicyeval}, mitigating this degradation in rollout quality is imperative.  A fundamental obstacle lies in the domain misalignment between low-dimensional control signals and high-dimensional video synthesis. Standard practices typically inject raw joint-space parameters, end-effector vectors, or abstract tokens as conditionings \citep{wu2024ivideogpt,rigter2024avid,zhu2025irasim,qu2025spatialvla}. While computationally compact, these representations are heavily tied to specific robot embodiments and carry limited spatial context \citep{zheng2025uniact}. They force the video generator to implicitly deduce cross-domain kinematics, frequently resulting in a failure to accurately render robot geometry or capture nuanced physical interactions. Furthermore, conventional architectures often lack an effective interaction mechanism between the action and video pathways, causing the generator to overlook fine-grained interaction information between the robot and objects.

\begin{figure}[t]
  \centering
  \includegraphics[width=0.95\linewidth]{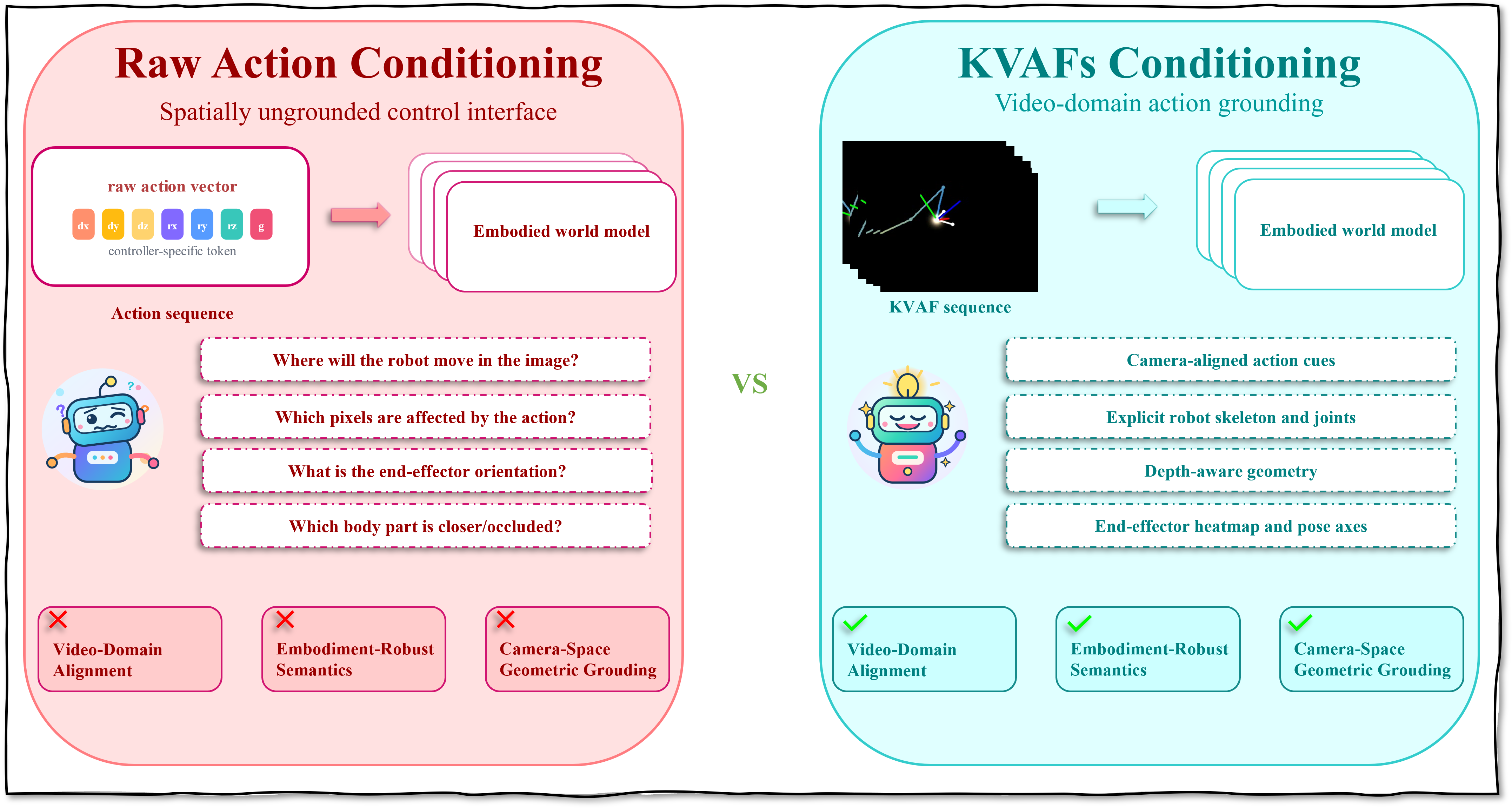}
  \caption{Comparison between direct low-dimensional action conditioning and the proposed \textit{Structured Kinematic-to-Visual Action Fields} (KVAFs).}
  \label{fig:kvaf_domain_alignment}
\end{figure}

To address these challenges, we propose \textbf{EA-WM}, an event-aware robotic video world model conditioned on \textit{Structured Kinematic-to-Visual Action Fields} (KVAFs). At the representation level, EA-WM constructs KVAFs by lifting low-dimensional robot actions and kinematic states into the target camera view through robot kinematics and camera projection. Specifically, KVAFs render depth-aware arm structures, joint landmarks, gripper geometry, end-effector heatmaps, and pose cues as temporally aligned visual fields. Compared with raw action vectors, KVAFs bring action conditioning into the same image domain as future video generation, thereby alleviating domain misalignment and providing richer spatial and geometric cues for robotic world modeling, as shown in Figure~\ref{fig:kvaf_domain_alignment}. At the architectural level, EA-WM augments a diffusion-transformer backbone with a dedicated KVAF branch and interval-based bidirectional fusion, facilitating continuous interaction between structured kinematic cues and video features. Crucially, we introduce an event-aware bidirectional fusion mechanism driven by \textit{Event-Difference Latent Supervision} (EDLS). By supervising event predictions with VAE-encoded temporal difference latents, EDLS compels the model to dynamically allocate attention not only to the robot's geometric progression but also to regions undergoing state transitions and interaction dynamics, leading to significantly more physically consistent rollouts. The key contributions of this work are summarized as follows:
\begin{itemize}
    \item We introduce \textbf{EA-WM}, an innovative Event-Aware Generative World Model framework for robotic video generation. By projecting low-dimensional kinematic states directly into the target camera view, our framework creates \textit{Structured Kinematic-to-Visual Action Fields} (KVAFs), fundamentally resolving the domain misalignment between raw action tokens and video-domain generation.

    \item We propose an \emph{event-aware bidirectional fusion} mechanism driven by \emph{Event-Difference Latent Supervision} (EDLS). This module explicitly regularizes cross-stream attention using temporal difference latents, compelling the model to dynamically attend to critical regions of robot motion, object state transitions, and robot-object interaction dynamics, leading to more physically consistent robot videos.

    \item We achieve superior performance on the  WorldArena benchmark. Compared to existing state-of-the-art baselines, EA-WM significantly enhances the physical adherence, 3D geometric accuracy, and fine-grained controllability of generated rollouts, establishing a new standard for robotic video modeling.
\end{itemize}

\section{Related Work}

\subsection{Robotic world models based on video generation models}
Recent diffusion-based video generators, including Stable Video Diffusion \citep{blattmann2023stablevideodiffusion}, Lumiere \citep{bartal2024lumiere}, CogVideoX, and the Wan series \citep{yang2025cogvideox,wanteam2025wan,wan2025wan22,gao2025wans2v}, have substantially improved temporal coherence, motion realism, and scene-consistent synthesis, making them increasingly useful as priors for world simulation \citep{openai2024sora,agarwal2025cosmos}. These priors are especially useful in robotics, where world models must predict how actions change future visual states. Existing robotic video world models span action-conditioned simulators and forward models for planning, model-based RL, and interaction simulation \citep{wu2024ivideogpt,rigter2024avid,zhu2025irasim}, infrastructure methods that use generated videos as synthetic trajectories, supervision, or policy-learning proxies \citep{jang2025dreamgen,liang2025videopolicy}, and learned video environments for scalable VLA evaluation and planning \citep{quevedo2025worldgym,li2025worldeval,tseng2025scalablepolicyeval}. Recent world-action models further couple video and action generation in unified frameworks \citep{zhu2025uwm,ye2026dreamzero,kim2026cosmospolicy}. Despite these advances, how to fully exploit action information to improve future robot-video generation remains insufficiently explored.

\subsection{World action models}
World Action Models (WAMs) are unified generative models that predict both robot actions and future visual states in an aligned manner, as formulated in DreamZero \citep{ye2026dreamzero}. Recent work \citep{zhu2025uwm,li2025uva,ye2026dreamzero,shen2025videovla,routray2025vipra,su2026wog,kim2026cosmospolicy,bi2025motus,pai2025mimicvideo} in this line mainly treats future video modeling as dense supervision or latent guidance for better action generation and policy learning. These WAM-style methods demonstrate that future videos can regularize action learning with physical dynamics, provide denser supervision than action-only imitation, and improve generalization for robot control. However, their emphasis remains largely on how video generation can improve action prediction or policy learning. The reverse direction---using action information as structured guidance to improve future robot-video generation---remains much less explored, even though rollout quality directly affects the usefulness of robotic world models as predictive simulators, planning substrates, and policy-evaluation environments \citep{jang2025dreamgen,liang2025videopolicy,quevedo2025worldgym,li2025worldeval,tseng2025scalablepolicyeval}.


A key question is therefore how action information should be represented. Existing methods have explored raw joint-space or end-effector vectors, 7-DoF control tokens, latent actions, universal action tokens, and spatial action abstractions. VideoVLA \citep{shen2025videovla}, UVA \citep{li2025uva}, Motus \citep{bi2025motus}, and ViPRA \citep{routray2025vipra} represent action information through numerical, latent, or video-conditioned action spaces, while UniAct \citep{zheng2025uniact} and SpatialVLA \citep{qu2025spatialvla} study more transferable or spatially structured action representations. While compact and effective for policy learning, these representations remain weakly grounded in the target camera view and carry limited explicit spatial and geometric information for video generation. WoG \citep{su2026wog} also highlights the challenge of preserving sufficient fine-grained information under compact future-conditioning representations. Recent visual representations lift actions into image or video space via RGB action targets \citep{shridhar2024genima}, virtual robot renders \citep{vosylius2024renderdiffuse}, pixel-grounded multiview action videos \citep{zhen2026actionimages}, or multiview heatmap videos \citep{li2026mvvdp}. However, their action channels are often intentionally compact and robot-centric: they prioritize joint targets, rendered robot bodies, arm-motion cues, or heatmaps, so fine-grained kinematic detail may still be limited, and robot-object interaction information is often modeled only indirectly through companion RGB branches or downstream video generation rather than encoded explicitly in the action representation itself.

\section{Method}
\label{method}
\begin{figure}[t]
  \centering
  \includegraphics[width=0.98\linewidth]{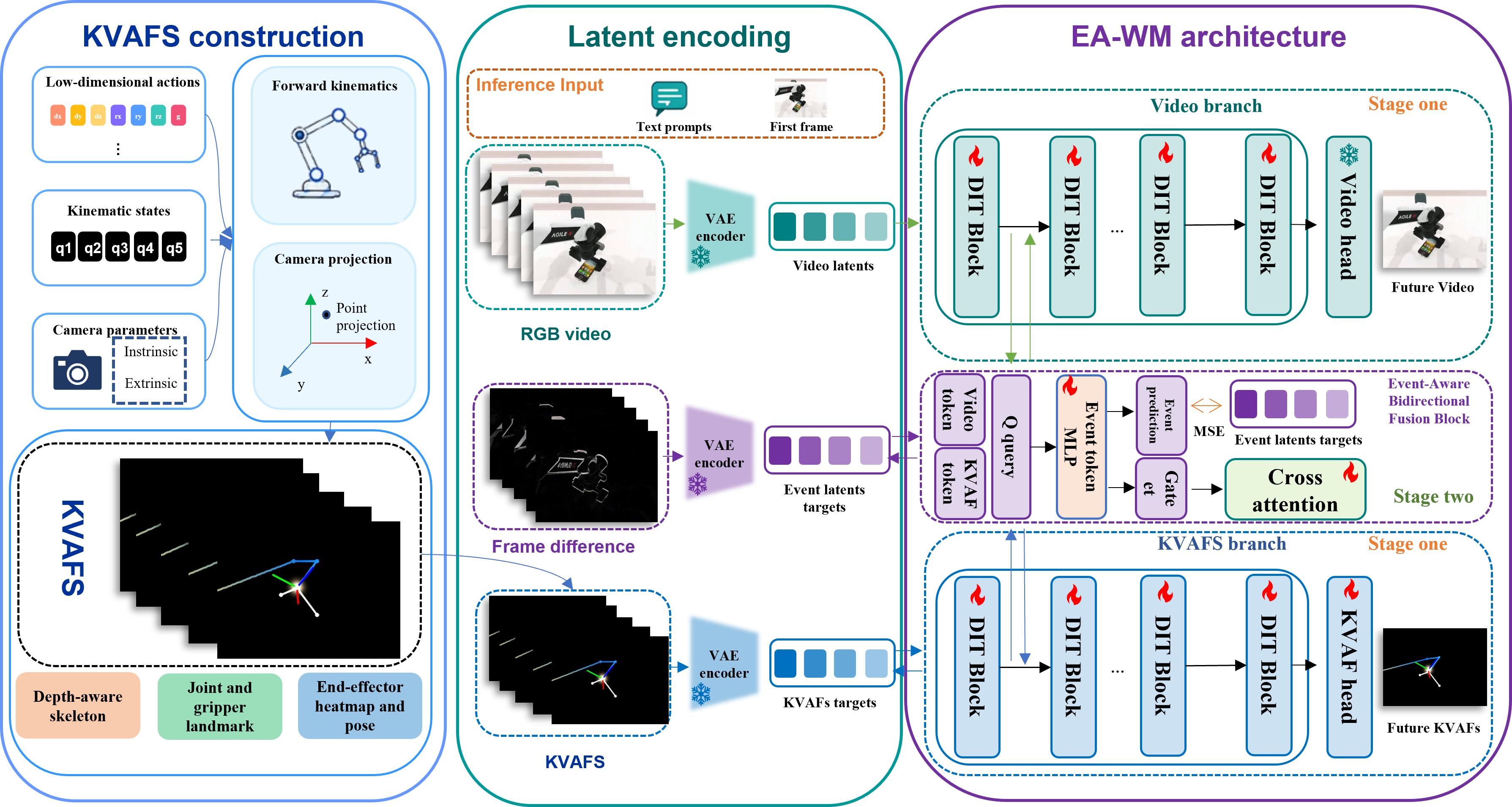}
  \caption{Overview of EA-WM. Robot actions and kinematic states are first lifted into camera-aligned KVAFs. RGB videos and KVAFs are encoded into the shared Wan2.2 latent space and processed by a video branch and a KVAF branch. Sparse event-aware bidirectional fusion exchanges information between the two branches, while EDLS guides the model to attend to motion and interaction changes.}
  \label{fig:eawm_pipeline}
\end{figure}

EA-WM is an action-guided robotic video world model built upon the Wan2.2-TI2V backbone \citep{wanteam2025wan}. The overall pipeline is shown in Figure~\ref{fig:eawm_pipeline}. Given an initial RGB observation, a language instruction, and a sequence of robot actions, our goal is to generate future robot videos that are not only visually plausible but also consistent with the commanded robot motion and robot-object interactions. Instead of injecting low-dimensional actions directly into the video generator, EA-WM first transforms actions and kinematic states into \textit{Structured Kinematic-to-Visual Action Fields} (KVAFs), which provide camera-aligned visual cues of robot geometry, motion, and end-effector intent. Both RGB videos and KVAFs are encoded into the latent space of the pretrained video VAE, allowing action information to guide generation in the same representation space as the target videos. To effectively exploit this structured action signal, EA-WM introduces a dedicated KVAF branch and event-aware bidirectional fusion blocks on top of the Wan2.2 video DiT. The KVAF branch learns to accurately predict KVAFs, encouraging the branch to understand the geometric and motion information embedded in KVAFs. Meanwhile, the event-aware bidirectional fusion blocks are guided by \emph{Event-Difference Latent Supervision} (EDLS), which uses VAE-encoded temporal difference latents to supervise event predictions, encouraging the model to capture object motion and robot-object interaction cues. Through cross-attention, the video backbone exchanges information with the KVAF branch, allowing structured action geometry and event-aware interaction cues to be shared with the generative stream for more accurate future video generation.

\subsection{Structured Kinematic-to-Visual Action Fields}

We introduce \emph{Structured Kinematic-to-Visual Action Fields} (KVAFs), which convert low-dimensional robot actions and kinematic states into camera-aligned visual fields for future video generation. Given a trajectory of length $T$, we construct a sequence $\mathcal{V}=\{\mathbf{V}_t\}_{t=1}^{T}$, where each frame $\mathbf{V}_t\in\mathbb{R}^{H\times W\times 3}$ is rendered in the target camera view. Here $H$ and $W$ denote the image height and width, and $t$ indexes the video timestep.

At time step $t$, we take arm joint values $\mathbf{q}_t$, gripper states $g_t$, end-effector poses $\boldsymbol{\xi}_t$, and camera parameters $(\mathbf{K}_t,\mathbf{E}_t)$ as input. Here $\mathbf{K}_t$ denotes the camera intrinsic matrix, and $\mathbf{E}_t$ denotes the world-to-camera extrinsic matrix. For simplicity, we omit the left/right arm index in the notation; in practice, KVAFs are constructed for both arms. We first recover 3D robot geometry by forward kinematics,
\begin{equation}
\label{eq:forward_kinematics}
\mathbf{T}^{W}_{k}(t)=\mathbf{T}^{W}_{k-1}(t)\mathbf{T}^{\mathrm{orig}}_{k}\mathbf{T}^{\mathrm{mot}}_{k}(q_{t,k}),
\end{equation}
where $\mathbf{T}^{W}_{k}(t)$ is the world-frame pose of the $k$-th robot link, $\mathbf{T}^{\mathrm{orig}}_{k}$ is the static joint transform from the robot kinematic chain, and $\mathbf{T}^{\mathrm{mot}}_{k}(q_{t,k})$ is the motion transform induced by the joint value $q_{t,k}$. The resulting link poses provide arm and gripper keypoints $\mathcal{P}_t=\{\mathbf{p}^{W}_{k}(t)\}$ in the world frame.

We then project these 3D keypoints into the camera plane by
\begin{equation}
\label{eq:camera_projection}
\mathbf{p}^{C}=\mathbf{E}_t
\begin{bmatrix}
\mathbf{p}^{W}\\1
\end{bmatrix},
\qquad
\hat{\mathbf{u}}=\mathbf{K}_t\mathbf{p}^{C}_{1:3},
\qquad
(u,v)=\left(\frac{\hat{u}_x}{\hat{u}_z},\frac{\hat{u}_y}{\hat{u}_z}\right).
\end{equation}
Here $\mathbf{p}^{W}$ and $\mathbf{p}^{C}$ denote the same robot keypoint in the world and camera coordinate frames, respectively; $\hat{\mathbf{u}}$ is the homogeneous image coordinate, and $(u,v)$ is the final projected pixel location. Points behind the camera are discarded.

Finally, we rasterize depth-aware arm skeletons, joint landmarks, gripper geometry, end-effector heatmaps, and pose axes on a black canvas to obtain
\begin{equation}
\label{eq:kvaf_rendering}
\mathbf{V}_t=\mathrm{Render}\big(\mathcal{P}_t,\boldsymbol{\xi}_t,\mathbf{K}_t,\mathbf{E}_t\big).
\end{equation}
The rendering operator composes the projected robot keypoints, end-effector pose cues, and gripper-related visual elements into an RGB visual action field. By expressing robot actions in the same image domain as the target rollout, KVAFs provide spatially grounded and temporally aligned motion cues that are more compatible with video-based robotic world modeling than raw action tokens.

\subsection{Event-Aware Generative World Model}

EA-WM builds upon the Wan2.2-TI2V backbone and preserves its original text-conditioned video denoising path. To inject structured action information, we encode KVAFs with the same video VAE and introduce a dedicated KVAF branch in the latent space. After patchification, the noised video latent and KVAF latent are represented as two token streams, denoted by $\mathbf{H}^{v}$ and $\mathbf{H}^{k}$. The video stream follows the original Wan2.2 DiT blocks, while the KVAF stream is modeled by a full-depth copy of these blocks, allowing action information to remain as a structured visual stream rather than being compressed into a low-dimensional token.

We insert event-aware fusion modules at a sparse set of layers $\mathcal{S}$. For $\ell\in\mathcal{S}$, an event MLP first computes a shared event representation from the current video and KVAF tokens, and then predicts both a per-token event gate and an event latent:
\begin{equation}
\label{eq:event_predictor}
\mathbf{M}_{\ell}
=
\Phi_{\ell}
\left(
\mathbf{H}^{v}_{\ell-1},
\mathbf{H}^{k}_{\ell}
\right),
\qquad
\mathbf{G}_{\ell}
=
\Gamma_{\ell}(\mathbf{M}_{\ell}),
\qquad
\hat{\mathbf{E}}_{\ell}
=
\Psi_{\ell}(\mathbf{M}_{\ell}).
\end{equation}
Here $\mathbf{G}_{\ell}$ is a per-token event gate that modulates cross-stream information exchange, while $\hat{\mathbf{E}}_{\ell}$ is the predicted event latent supervised by EDLS. Since both outputs are derived from the shared event representation $\mathbf{M}_{\ell}$, EDLS encourages $\mathbf{M}_{\ell}$ to encode temporal changes and interaction cues, which are then passed to the gate $\mathbf{G}_{\ell}$ and used to regulate video--KVAF fusion.

The event gate controls bidirectional cross-attention between the two streams:
\begin{equation}
\label{eq:video_cross_attention}
\tilde{\mathbf{H}}^{v}_{\ell-1}
=
\mathbf{H}^{v}_{\ell-1}
+
\mathbf{G}_{\ell}\odot
\mathrm{CA}_{v\leftarrow k}
\left(
\mathbf{H}^{v}_{\ell-1},
\mathbf{H}^{k}_{\ell}
\right),
\end{equation}
\begin{equation}
\label{eq:kvaf_cross_attention}
\tilde{\mathbf{H}}^{k}_{\ell}
=
\mathbf{H}^{k}_{\ell}
+
\mathbf{G}_{\ell}\odot
\mathrm{CA}_{k\leftarrow v}
\left(
\mathbf{H}^{k}_{\ell},
\mathbf{H}^{v}_{\ell-1}
\right).
\end{equation}
The updated video tokens are then processed by the original Wan2.2 block, and the updated KVAF tokens are passed to the next KVAF branch block. At non-fusion layers, the two streams are updated without cross-stream fusion. In this way, EDLS does not merely add an auxiliary event prediction head; it shapes the shared event representation that produces the gate, and the gate directly controls how much KVAF information is injected into the video stream and how much scene information is fed back to the KVAF stream.

To construct EDLS, we compute a frame-difference video from the input sequence $\{I_{\tau}\}_{\tau=1}^{T}$:
\begin{equation}
\label{eq:frame_difference}
\Delta I_1=\mathbf{0},
\qquad
\Delta I_{\tau}=|I_{\tau}-I_{\tau-1}|,\quad \tau=2,\dots,T,
\end{equation}
and encode it with the same VAE to obtain the event latent target $\mathbf{E}$. Let $\hat{\mathbf{Y}}^{v}$ and $\hat{\mathbf{Y}}^{k}$ be the predictions of the video and KVAF heads, and $\mathbf{Y}^{v}$ and $\mathbf{Y}^{k}$ be their corresponding flow-matching targets. The total training objective is
\begin{equation}
\label{eq:training_objective}
\mathcal{L}
=
\omega(t)
\left(
\left\|
\hat{\mathbf{Y}}^{v}-\mathbf{Y}^{v}
\right\|_2^2
+
\left\|
\hat{\mathbf{Y}}^{k}-\mathbf{Y}^{k}
\right\|_2^2
\right)
+
\lambda_{\mathrm{evt}}
\frac{1}{|\mathcal{S}|}
\sum_{\ell\in\mathcal{S}}
\left\|
\mathrm{Unpatchify}(\hat{\mathbf{E}}_{\ell})-\mathbf{E}
\right\|_2^2 .
\end{equation}
Here $\omega(t)$ denotes the timestep-dependent flow-matching training weight used by the scheduler, and $\lambda_{\mathrm{evt}}$ controls the strength of EDLS. The first video frame is treated as the image condition and excluded from the video prediction loss. This objective encourages EA-WM to use KVAF information in regions where motion and interaction changes occur, preserving action-consistent robot geometry while remaining sensitive to object motion and robot-object interactions.

\section{Experiments}
\label{experiments}
\subsection{Experimental Details}
 We train EA-WM in two stages: the first stage freezes the event-aware fusion modules and trains the main DiT LoRA, KVAF-branch LoRA, and KVAF head to stabilize the two streams; the second stage unfreezes the fusion modules to learn cross-stream event-aware interaction. We evaluate generated videos with WorldArena \citep{shang2026worldarena}, a unified benchmark for embodied world models. We use a LoRA rank of 32, and set the learning rate to $8\times10^{-5}$ for all trainable components. Training is conducted on 32 H100 GPUs with a batch size of 32. Following WorldArena, we use the same Robotwin dataset and segmentation method and select three dimensions for evaluation: Physics Adherence, 3D Accuracy, and Controllability. These dimensions measure interaction plausibility, trajectory consistency, depth and perspective accuracy, instruction following, and semantic consistency. We compare EA-WM with multiple representative video world models under the selected metrics.

\subsection{Experimental Results}
\paragraph{Quantitative Analysis.}
Following WorldArena, we evaluate EA-WM on three dimensions: Physics Adherence, 3D Accuracy, and Controllability. All metrics are normalized to $[0,1]$, and higher is better. We report $\mathrm{P3CScore}$, the average score over the six selected metrics multiplied by 100. The final results are shown in Table~\ref{tab:worldarena_p3c}.

\begin{table}[t]
\centering
\caption{
Quantitative comparison on selected WorldArena metrics.All metrics are normalized to $[0,1]$ and higher is better.
$\mathrm{P3CScore}$ is the average over the six selected metrics multiplied by 100.
}
\label{tab:worldarena_p3c}
\setlength{\tabcolsep}{4pt}
\resizebox{\linewidth}{!}{
\begin{tabular}{lccccccc}
\toprule
 & \multicolumn{2}{c}{Physics Adherence}
 & \multicolumn{2}{c}{3D Accuracy}
 & \multicolumn{2}{c}{Controllability}
 & \multirow{2}{*}{$\mathrm{P3CScore}$} \\
\cmidrule(lr){2-3}
\cmidrule(lr){4-5}
\cmidrule(lr){6-7}
Model
& \shortstack{Interaction\\Quality}
& \shortstack{Trajectory\\Accuracy}
& \shortstack{Depth\\Accuracy}
& Perspectivity
& \shortstack{Instruction\\Following}
& \shortstack{Semantic\\Alignment}
&  \\
\midrule
CogVideoX
& 0.594 & 0.353
& 0.910 & 0.783
& 0.727 & \textbf{0.898}
& 71.08 \\

WoW
& 0.556 & 0.206
& 0.728 & 0.767
& 0.569 & 0.884
& 61.83 \\

Wan 2.2
& 0.518 & 0.163
& 0.777 & 0.766
& 0.538 & 0.888
& 60.83 \\

TesserAct
& 0.580 & 0.140
& 0.716 & 0.792
& 0.615 & 0.878
& 62.02 \\

GigaWorld-0
& 0.537 & 0.155
& 0.632 & 0.760
& 0.616 & 0.859
& 59.32 \\

Vidar
& 0.535 & 0.193
& 0.787 & 0.759
& 0.591 & 0.883
& 62.47 \\

Cosmos-Predict 2.5 (text)
& 0.387 & 0.082
& 0.705 & 0.796
& 0.266 & 0.773
& 50.15 \\

Genie Envisioner
& 0.205 & 0.068
& 0.866 & 0.528
& 0.203 & 0.854
& 45.40 \\

\midrule
\textbf{EA-WM (Ours)}
& \textbf{0.682} & \textbf{0.430}
& \textbf{0.959} & \textbf{0.838}
& \textbf{0.792} & 0.895
& \textbf{76.60} \\
\bottomrule
\end{tabular}
}
\end{table}

EA-WM achieves the best overall $\mathrm{P3CScore}$ of 76.60, outperforming the strongest baseline, CogVideoX, by 5.52 points. The improvement is particularly clear on metrics related to action-induced motion and interaction: EA-WM improves Interaction Quality from 0.594 to 0.682, Trajectory Accuracy from 0.353 to 0.430, and Instruction Following from 0.727 to 0.792. These gains suggest that introducing KVAFs and event-aware bidirectional fusion blocks helps the model better preserve robot-object interaction, follow commanded motion trajectories, and generate videos that are more consistent with the task instruction. EA-WM also improves Depth Accuracy and Perspectivity by 0.049 and 0.042 over the strongest baselines, respectively, indicating stronger 3D consistency. Overall, EA-WM outperforms all compared models on five out of six selected metrics, demonstrating stronger physics adherence, 3D consistency, and controllability for robotic video world modeling.

\begin{figure}[t]
  \centering
  \includegraphics[width=0.98\linewidth]{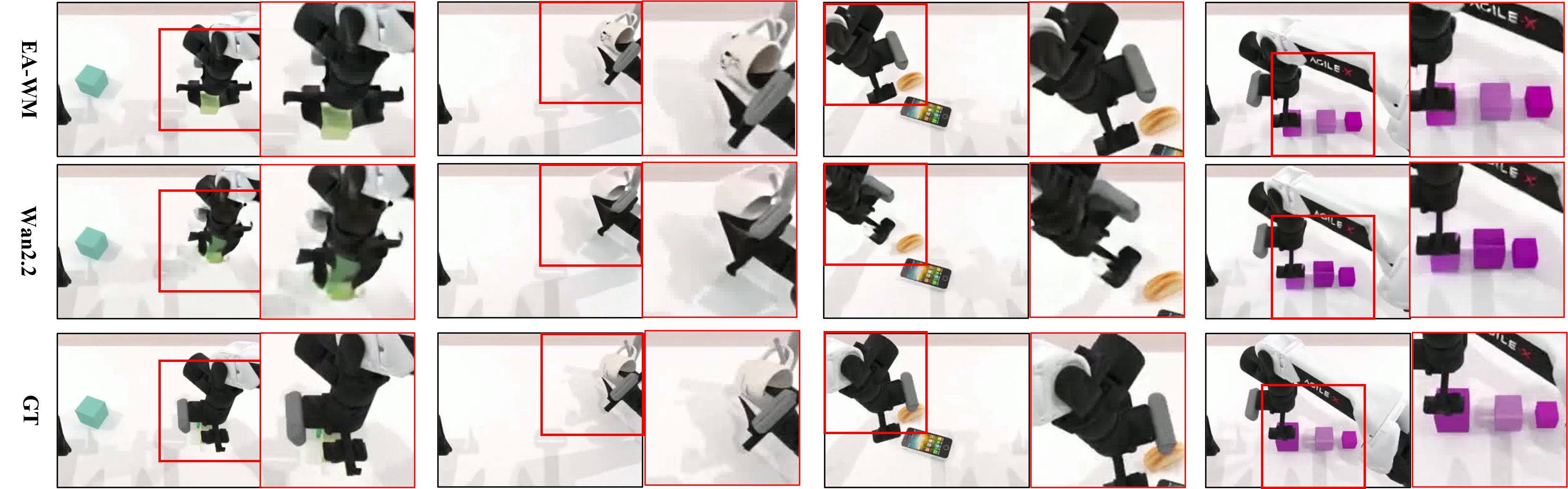}
  \caption{
  Qualitative comparison on four randomly selected RoboTwin 2.0 tasks.
  For each task, we visualize results at the same temporal step.
  Red boxes highlight local robot-object interaction regions.
  }
  \label{fig:qualitative_results}
\end{figure}

\paragraph{Qualitative Analysis.}
We further conduct qualitative comparisons on four randomly selected tasks, visualizing GT, the Wan2.2 baseline, and EA-WM at the same temporal step for each task. As shown in Figure~\ref{fig:qualitative_results}, we compare the results in four aspects: robot-object interaction, preservation of object geometry, physical consistency of robots, and completion of tasks. Compared with the Wan2.2 baseline, EA-WM jointly models KVAF and video streams through event-aware bidirectional fusion, allowing the model to better follow action-induced motion cues and focus on regions where object states and robot-object interactions change. As a result, EA-WM generates more accurate gripper-object contact relations, better preserves object identity and local geometry, and maintains more physically plausible robot poses during manipulation. Overall, EA-WM is closer to GT in both interaction fidelity and task progress, demonstrating stronger action-consistent future video generation.

\subsection{Ablation Study}

To verify the effectiveness of each component in EA-WM, we conduct ablation studies on the same selected WorldArena metrics used in quantitative evaluation. We compare the full model with three variants: the original Wan2.2 backbone, a variant that replaces KVAFs with numerical action representations, and a variant that removes event-aware fusion while keeping bidirectional cross-attention for video--KVAF interaction. We report Physics Adherence, 3D Accuracy, Controllability, and the averaged $\mathrm{P3CScore}$.

The ablation results reveal the complementary roles of KVAFs and EAF. Compared with the variant without KVAFs, the full EA-WM improves $\mathrm{P3CScore}$ by 5.63 points, with particularly clear gains on Trajectory Accuracy and Depth Accuracy. This suggests that camera-aligned KVAFs provide useful geometric motion cues beyond numerical action inputs, especially for preserving robot trajectories and 3D structure in the generated videos. Compared with the variant without EAF, EA-WM improves $\mathrm{P3CScore}$ by 1.80 points and shows notable gains on Interaction Quality and Perspectivity, indicating that EDLS-guided event-aware fusion helps the model better focus on motion changes and robot-object interaction regions during cross-stream information exchange. Although the variant without EAF slightly outperforms EA-WM on Semantic Alignment, the full model achieves the best overall score and more balanced performance across physics adherence, 3D accuracy, and controllability.

\begin{figure*}[t]
  \centering
  \includegraphics[width=0.98\textwidth]{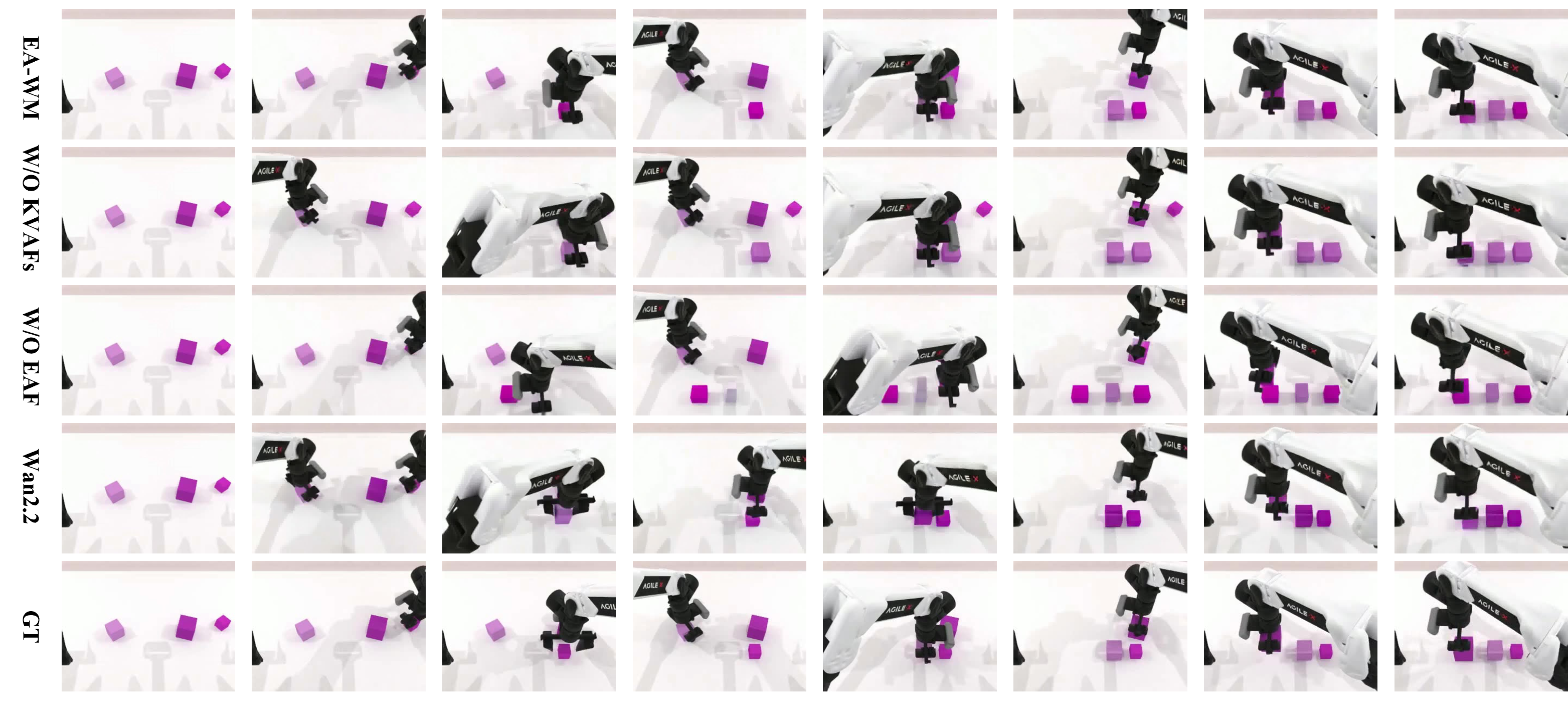}
  \caption{Ablation visualization on the \texttt{ranking\_block\_size} task. }
  \label{fig:eawm_sequence}
\end{figure*}

We further visualize the \texttt{ranking\_block\_size} task to more intuitively compare the differences among ablation variants. As shown in Figure~\ref{fig:eawm_sequence}, the two ablations exhibit complementary failure modes. The variant without KVAFs can better maintain the local relations between the robot arm and objects, but it is less accurate in modeling the grasping trajectory and action-induced motion. In contrast, the variant without event-aware fusion can better infer the grasping route of the robot arm from KVAFs, but it struggles to preserve object consistency, such as the relative shape, size, and spatial arrangement of the blocks. By combining KVAFs with event-aware fusion, EA-WM handles both aspects more effectively: it follows the spatial motion pattern of the robot while preserving object consistency and robot-object interaction fidelity, generating videos that better match the physical progression of the task.

\begin{table}[t]
\centering
\caption{
Ablation study on selected WorldArena metrics.
All metrics are normalized to $[0,1]$ and higher is better.
$\mathrm{P3CScore}$ is the average over the six selected metrics multiplied by 100.
}
\label{tab:ablation_study}
\setlength{\tabcolsep}{4pt}
\resizebox{\linewidth}{!}{
\begin{tabular}{lccccccc}
\toprule
 & \multicolumn{2}{c}{Physics Adherence}
 & \multicolumn{2}{c}{3D Accuracy}
 & \multicolumn{2}{c}{Controllability}
 & \multirow{2}{*}{$\mathrm{P3CScore}$} \\
\cmidrule(lr){2-3}
\cmidrule(lr){4-5}
\cmidrule(lr){6-7}
Variant
& \shortstack{Interaction\\Quality}
& \shortstack{Trajectory\\Accuracy}
& \shortstack{Depth\\Accuracy}
& Perspectivity
& \shortstack{Instruction\\Following}
& \shortstack{Semantic\\Alignment}
&  \\
\midrule
Wan2.2
& 0.518 & 0.163
& 0.777 & 0.766
& 0.538 & 0.888
& 60.83 \\

w/o KVAFs
& 0.670 & 0.298
& 0.884 & 0.791
& 0.726 & 0.889
& 70.97 \\

w/o EAF
& 0.656 & 0.407
& 0.947 & 0.800
& 0.782 & \textbf{0.896}
& 74.80 \\

\textbf{EA-WM}
& \textbf{0.682} & \textbf{0.430}
& \textbf{0.959} & \textbf{0.838}
& \textbf{0.792} & 0.895
& \textbf{76.60} \\
\bottomrule
\end{tabular}
}
\end{table}

\subsection{Additional Analysis}
\begin{figure*}[t]
\centering
\begin{minipage}[t]{0.40\textwidth}
\vspace{0pt}
\centering
\captionof{table}{
Numerical action recovery from generated KVAFs.
Errors are reported for translation, rotation, and gripper actions; lower is better.
}
\label{tab:kvaf_action_recovery}
\vspace{0.4em}
\small
\setlength{\tabcolsep}{2pt}
{\renewcommand{\arraystretch}{1.7}
\resizebox{\linewidth}{!}{%
\begin{tabular}{@{}l@{\hspace{1.0em}}c@{\hspace{1.0em}}c@{\hspace{1.0em}}c@{}}
\toprule
Method 
& \shortstack{Translation\\Error}
& \shortstack{Rotation\\Error}
& \shortstack{Gripper\\Error} \\
\midrule
Raw-action baseline
& \textbf{0.004} & \textbf{0.009} & \textbf{0.013} \\
KVAF recovery
& 0.0155 & 0.110 & 0.039 \\
\bottomrule
\end{tabular}
}
}

\vspace{0.5em}
\footnotesize
Heuristic KVAF recovery detection rate: $\sim 0.45$.
\end{minipage}
\hfill
\begin{minipage}[t]{0.56\textwidth}
\vspace{0pt}
\centering
\includegraphics[width=\linewidth]{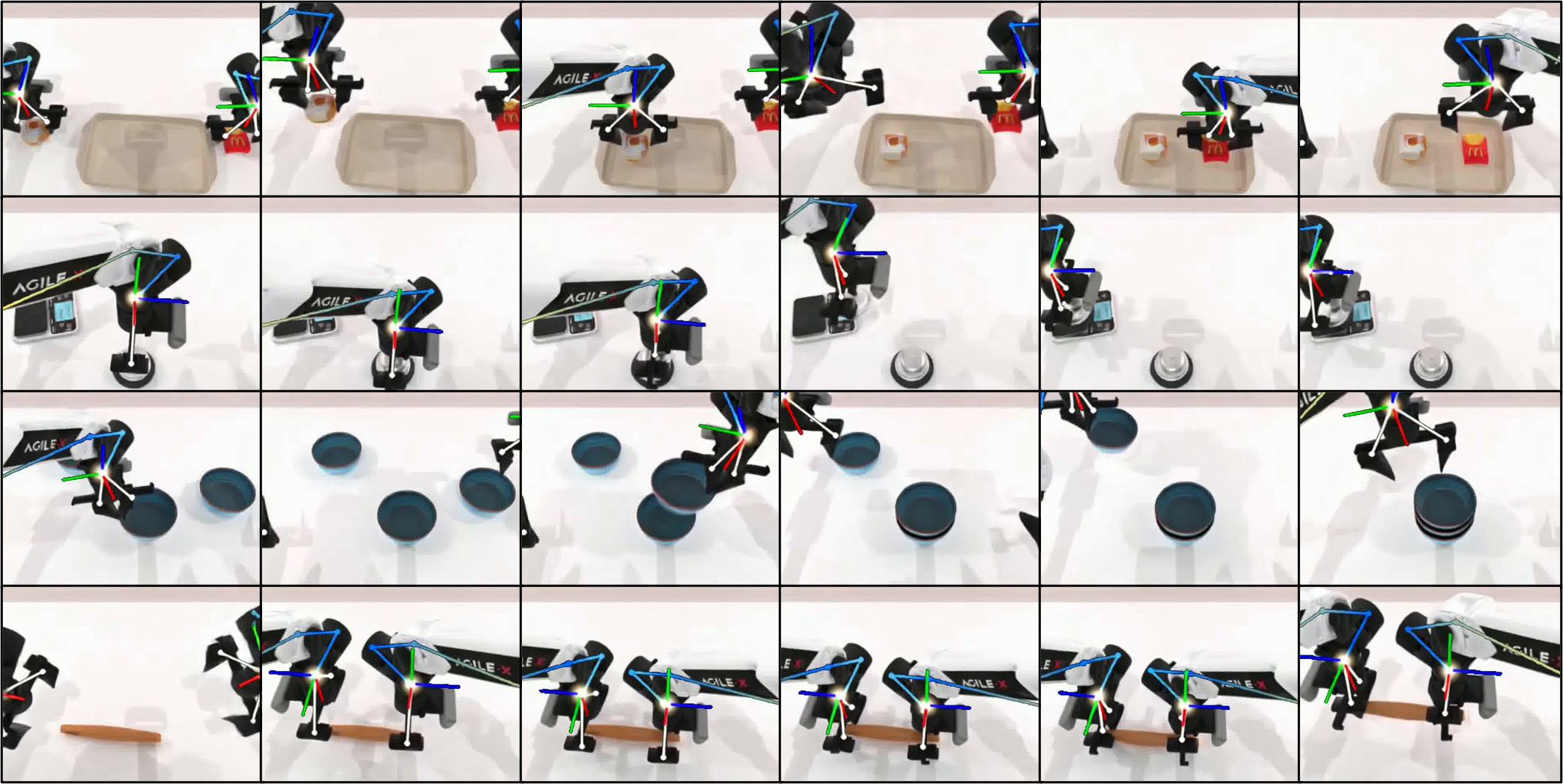}
\vspace{-0.4em}
\captionof{figure}{
KVAFs overlays on generated videos. The overlays align with generated robot motion and reveal interpretable spatial cues learned from KVAFs.
}
\label{fig:kvaf_overlay_visualization}
\end{minipage}
\end{figure*}

\paragraph{Action recovery and image-domain alignment of KVAFs.}
To better understand the information encoded in KVAFs, we conduct a preliminary action-recovery analysis in the numerical action space. We use a heuristic procedure to recover end-effector translation, orientation, and gripper states from generated KVAF videos, and compare the recovered actions with predictions from a raw-action baseline that directly predicts numerical actions. As shown in Table~\ref{tab:kvaf_action_recovery}, numerical action prediction still performs better on translation, rotation, and gripper errors. This gap is expected, since KVAFs are designed as image-domain action fields rather than direct numerical action outputs, and our current recovery pipeline depends on detecting small visual cues such as heatmap centers, pose axes, and gripper geometry. The detection rate of the heuristic KVAF recovery is approximately $0.45$, suggesting that mapping structured visual action fields back to precise numerical actions remains a challenging problem. Nevertheless, the recovery results indicate that KVAFs contain partially recoverable action information, and motivate future work on more reliable bridges between image-domain action fields and numerical action spaces. More importantly, the overlay visualizations in Figure~\ref{fig:kvaf_overlay_visualization} show that the predicted KVAFs are well aligned with the generated robot motion across randomly selected tasks. The projected skeletons, end-effector cues, and pose axes follow the robot configuration in the generated videos, indicating that the network has learned the spatial and physical information contained in KVAFs rather than treating them as abstract conditioning signals. This image-domain alignment helps connect action-induced robot motion with object locations and local interaction regions, providing an interpretable bridge between low-dimensional actions and future video generation.

\paragraph{KVAF-conditioned video generation.}
We further compare KVAF-conditioned generation with mainstream action- or trajectory-conditioned baselines. This analysis focuses on whether representing actions as camera-aligned visual fields can provide a stronger conditioning interface than numerical actions or trajectories. As shown in Table~\ref{tab:kvaf_condition_action_models}, KVAF-conditioned generation achieves the best performance across all six selected metrics and obtains the highest $\mathrm{P3CScore}$. Compared with raw action- or trajectory-conditioned models, KVAFs provide richer spatial and geometric cues in the video domain, leading to better physics adherence, 3D consistency, and controllability. These results further suggest that transforming action information into structured visual action fields is an effective way to guide robotic video world modeling.

\begin{table}[t]
\centering
\caption{
Comparison with action- or trajectory-conditioned baselines on selected WorldArena metrics.
All metrics are normalized to $[0,1]$ and higher is better.
$\mathrm{P3CScore}$ is the average over the six selected metrics multiplied by 100.
}
\label{tab:kvaf_condition_action_models}
\setlength{\tabcolsep}{4pt}
\resizebox{\linewidth}{!}{
\begin{tabular}{lccccccc}
\toprule
 & \multicolumn{2}{c}{Physics Adherence}
 & \multicolumn{2}{c}{3D Accuracy}
 & \multicolumn{2}{c}{Controllability}
 & \multirow{2}{*}{$\mathrm{P3CScore}$} \\
\cmidrule(lr){2-3}
\cmidrule(lr){4-5}
\cmidrule(lr){6-7}
Model
& \shortstack{Interaction\\Quality}
& \shortstack{Trajectory\\Accuracy}
& \shortstack{Depth\\Accuracy}
& Perspectivity
& \shortstack{Instruction\\Following}
& \shortstack{Semantic\\Alignment}
&  \\
\midrule
CtrlWorld
& 0.621 & 0.477
& 0.930 & 0.796
& 0.727 & 0.891
& 74.03 \\

IRASim
& 0.566 & 0.364
& 0.931 & 0.779
& 0.660 & 0.885
& 69.75 \\

Cosmos-Predict 2.5 (action)
& 0.550 & 0.295
& 0.886 & 0.764
& 0.584 & 0.888
& 66.12 \\

RoboMaster
& 0.536 & 0.116
& 0.834 & 0.759
& 0.577 & 0.876
& 61.63 \\

\midrule
\textbf{KVAF-conditioned generation}
& \textbf{0.682} & \textbf{0.494}
& \textbf{0.977} & \textbf{0.850}
& \textbf{0.784} & \textbf{0.901}
& \textbf{78.13} \\
\bottomrule
\end{tabular}
}
\end{table}

\section{Limitations}
\label{limitation}
EA-WM requires robot kinematics, camera calibration, and synchronized action-state logs to construct KVAFs. While this yields accurate, well-aligned action representations in simulation, real-world deployment may be affected by calibration errors, occlusions, sensor noise, or embodiment shifts. Current KVAFs mainly encode robot-side geometric motion cues, while object-state changes and robot-object interactions are captured indirectly through event-aware fusion and EDLS. Extending KVAFs with object-centric or contact-aware visual fields is a promising future direction.

\section{Conclusion}
We present \textbf{EA-WM}, an event-aware generative world model for action-consistent robot-video generation. Unlike current WAMs that mainly optimize action prediction or policy learning, EA-WM focuses on preserving robot spatial motion and robot-object interaction dynamics in generated videos. It maps actions and kinematic states into \textit{Structured Kinematic-to-Visual Action Fields} (KVAFs) aligned with the visual generation domain, and integrates them through EDLS-guided event-aware bidirectional fusion. On WorldArena, EA-WM achieves the best $\mathrm{P3CScore}$ among compared models.



\bibliographystyle{plainnat}
\bibliography{main}

\clearpage
\appendix
\raggedbottom

\section{Appendix}
\subsection{Detailed Network Architecture Pipeline}
\label{app:network_pipeline}

This section describes the full EA-WM network pipeline. We focus on how RGB videos, KVAFs, and event-difference targets are processed by the Wan2.2-based architecture. 

\paragraph{Latent preparation.}
Given an RGB video sequence $\mathbf{X}=\{I_\tau\}_{\tau=1}^{T}$, the corresponding KVAF sequence $\mathbf{K}$, and text condition $\mathbf{c}$, EA-WM first encodes both videos and KVAFs with the same pretrained video VAE:
\[
\mathbf{z}^{v}_{0}=\mathrm{VAE}(\mathbf{X}), 
\qquad 
\mathbf{z}^{k}_{0}=\mathrm{VAE}(\mathbf{K}).
\]
This places the target video and action-derived KVAFs in a shared latent space. During flow-matching training, both latents are perturbed at timestep $t$, producing noised latents $\mathbf{z}^{v}_{t}$ and $\mathbf{z}^{k}_{t}$ together with their training targets $\mathbf{Y}^{v}$ and $\mathbf{Y}^{k}$.

\paragraph{Dual-stream denoising.}
After patchification, the video latent and KVAF latent are represented as two token streams,
\[
\mathbf{H}^{v}_{0}=\mathcal{P}(\mathbf{z}^{v}_{t}), 
\qquad
\mathbf{H}^{k}_{0}=\mathcal{P}(\mathbf{z}^{k}_{t}).
\]
The video stream follows the original Wan2.2 DiT blocks, while the KVAF stream is processed by a full-depth copy of the DiT blocks. Event-aware fusion modules are inserted at a sparse set of layers $\mathcal{S}$, allowing the two streams to exchange information only at selected depths. This keeps the KVAF branch expressive while avoiding dense, overly expensive cross-stream interaction.

\paragraph{Event-Difference Latent Supervision.}
EA-WM uses frame-difference latents as event supervision. Given the input video, we compute
\[
\Delta I_1=\mathbf{0}, 
\qquad 
\Delta I_\tau=|I_\tau-I_{\tau-1}|,\quad \tau=2,\ldots,T,
\]
and encode the difference sequence with the same video VAE:
\[
\mathbf{E}=\mathrm{VAE}(\{\Delta I_\tau\}_{\tau=1}^{T}).
\]
This target supervises the event prediction branch in each fusion module. Since the event prediction and event gate share the same event representation, EDLS also shapes the gate that controls video--KVAF cross-attention.

\begin{algorithm}[H]
\caption{EA-WM training step}
\label{alg:eawm_training_compact}
\begin{algorithmic}[1]
\Require RGB video $\mathbf{X}$, KVAF sequence $\mathbf{K}$, text condition $\mathbf{c}$, fusion layers $\mathcal{S}$
\State Encode latents: $\mathbf{z}^{v}_{0}\gets \mathrm{VAE}(\mathbf{X})$, $\mathbf{z}^{k}_{0}\gets \mathrm{VAE}(\mathbf{K})$
\State Sample timestep $t$ and construct noised latents and flow targets:
\[
(\mathbf{z}^{v}_{t},\mathbf{Y}^{v}),\;(\mathbf{z}^{k}_{t},\mathbf{Y}^{k})
\gets \mathrm{FlowMatch}(\mathbf{z}^{v}_{0},\mathbf{z}^{k}_{0},t)
\]
\State Construct EDLS target: $\mathbf{E}\gets \mathrm{VAE}(\{|I_\tau-I_{\tau-1}|\}_{\tau=1}^{T})$
\State Patchify: $\mathbf{H}^{v}_{0}\gets \mathcal{P}(\mathbf{z}^{v}_{t})$, $\mathbf{H}^{k}_{0}\gets \mathcal{P}(\mathbf{z}^{k}_{t})$
\For{$\ell=1$ to $L$}
    \State Update KVAF stream: $\bar{\mathbf{H}}^{k}_{\ell}\gets \mathcal{B}_{\ell}(\mathbf{H}^{k}_{\ell-1},\mathbf{c},t)$
    \If{$\ell\in\mathcal{S}$}
        \State $\tilde{\mathbf{H}}^{v}_{\ell-1},\mathbf{H}^{k}_{\ell},\hat{\mathbf{E}}_{\ell}
        \gets \mathrm{EventFusion}_{\ell}(\mathbf{H}^{v}_{\ell-1},\bar{\mathbf{H}}^{k}_{\ell})$
    \Else
        \State $\tilde{\mathbf{H}}^{v}_{\ell-1}\gets\mathbf{H}^{v}_{\ell-1}$, \quad $\mathbf{H}^{k}_{\ell}\gets\bar{\mathbf{H}}^{k}_{\ell}$
    \EndIf
    \State Update video stream: $\mathbf{H}^{v}_{\ell}\gets\mathcal{D}_{\ell}(\tilde{\mathbf{H}}^{v}_{\ell-1},\mathbf{c},t)$
\EndFor
\State Predict $\hat{\mathbf{Y}}^{v}\gets \mathrm{Head}_{v}(\mathbf{H}^{v}_{L})$, \quad $\hat{\mathbf{Y}}^{k}\gets \mathrm{Head}_{k}(\mathbf{H}^{k}_{L})$
\State Compute $\mathcal{L}=\omega(t)(\|\hat{\mathbf{Y}}^{v}-\mathbf{Y}^{v}\|_2^2+\|\hat{\mathbf{Y}}^{k}-\mathbf{Y}^{k}\|_2^2)+\lambda_{\mathrm{evt}}\mathcal{L}_{\mathrm{EDLS}}$
\State \Return $\mathcal{L}$
\end{algorithmic}
\end{algorithm}

\paragraph{Event-aware bidirectional fusion.}
The fusion block is the key interaction module between the video stream and the KVAF stream. It first predicts a shared event representation, from which both the event gate and the event latent are derived. The event latent is supervised by EDLS, while the event gate modulates the strength of bidirectional cross-attention. Therefore, EDLS affects video generation not only through an auxiliary prediction loss, but also through the gate that controls information exchange between the two streams.

\begin{algorithm}[t]
\caption{Event-aware bidirectional fusion at layer $\ell$}
\label{alg:event_fusion_compact}
\begin{algorithmic}[1]
\Require Video tokens $\mathbf{H}^{v}_{\ell-1}$, KVAF tokens $\bar{\mathbf{H}}^{k}_{\ell}$
\State Predict event gate and event latent:
\[
\mathbf{G}_{\ell},\hat{\mathbf{E}}_{\ell}
\gets
\Phi_{\ell}(\mathbf{H}^{v}_{\ell-1},\bar{\mathbf{H}}^{k}_{\ell})
\]
\State Video reads KVAF:
\[
\mathbf{R}^{v}_{\ell}
\gets
\mathrm{CA}_{v\leftarrow k}(\mathbf{H}^{v}_{\ell-1},\bar{\mathbf{H}}^{k}_{\ell})
\]
\State KVAF reads video:
\[
\mathbf{R}^{k}_{\ell}
\gets
\mathrm{CA}_{k\leftarrow v}(\bar{\mathbf{H}}^{k}_{\ell},\mathbf{H}^{v}_{\ell-1})
\]
\State Apply event-gated residual fusion:
\[
\tilde{\mathbf{H}}^{v}_{\ell-1}
=
\mathbf{H}^{v}_{\ell-1}
+
\mathbf{G}_{\ell}\odot \mathbf{R}^{v}_{\ell}
\]
\[
\mathbf{H}^{k}_{\ell}
=
\bar{\mathbf{H}}^{k}_{\ell}
+
\mathbf{G}_{\ell}\odot \mathbf{R}^{k}_{\ell}
\]
\State \Return $\tilde{\mathbf{H}}^{v}_{\ell-1}$, $\mathbf{H}^{k}_{\ell}$, $\hat{\mathbf{E}}_{\ell}$
\end{algorithmic}
\end{algorithm}

\paragraph{Training objective.}
The full training loss consists of video flow matching, KVAF flow matching, and EDLS:
\[
\mathcal{L}
=
\omega(t)
\left(
\left\|
\hat{\mathbf{Y}}^{v}-\mathbf{Y}^{v}
\right\|_2^2
+
\left\|
\hat{\mathbf{Y}}^{k}-\mathbf{Y}^{k}
\right\|_2^2
\right)
+
\lambda_{\mathrm{evt}}
\frac{1}{|\mathcal{S}|}
\sum_{\ell\in\mathcal{S}}
\left\|
\mathrm{Unpatchify}(\hat{\mathbf{E}}_{\ell})-\mathbf{E}
\right\|_2^2 .
\]
The video loss trains the future video generation branch, the KVAF loss encourages the KVAF branch to preserve structured robot geometry, and EDLS guides the fusion modules to focus on motion and interaction changes.

\subsection{KVAF Construction}
\label{app:kvaf_construction}

This section describes how we construct \textit{Structured Kinematic-to-Visual Action Fields} (KVAFs) from robot actions and kinematic states. KVAFs are designed to convert low-dimensional robot-side information into camera-aligned visual action fields, so that action conditioning can be expressed in the same image domain as future video generation.

\paragraph{Inputs.}
For each episode, we assume access to the robot URDF, embodiment configuration, camera parameters, and per-frame robot states. At time step $t$, the input to KVAF construction is
\[
\mathcal{I}_t =
\left(
\mathbf{q}^{L}_t,\mathbf{q}^{R}_t,
g^{L}_t,g^{R}_t,
\boldsymbol{\xi}^{L}_t,\boldsymbol{\xi}^{R}_t,
\mathbf{K}_t,\mathbf{E}_t
\right),
\]
where $\mathbf{q}^{L}_t,\mathbf{q}^{R}_t$ are the left/right arm joint values, $g^{L}_t,g^{R}_t$ are normalized gripper states, $\boldsymbol{\xi}^{L}_t,\boldsymbol{\xi}^{R}_t$ are end-effector poses, and $(\mathbf{K}_t,\mathbf{E}_t)$ are camera intrinsics and extrinsics. Each end-effector pose is represented by a 3D position and a quaternion:
\[
\boldsymbol{\xi}^{a}_t=(\mathbf{p}^{a}_{t},\mathbf{q}^{a}_{t}), 
\qquad a\in\{L,R\}.
\]

\paragraph{Kinematic lifting.}
We first lift low-dimensional joint states into 3D robot geometry using forward kinematics. Let $\mathbf{T}^{W}_{a,k}(t)\in SE(3)$ denote the world-frame transform of the $k$-th link of arm $a$ at time $t$. Starting from the robot base transform, each link transform is computed recursively:
\[
\mathbf{T}^{W}_{a,k}(t)
=
\mathbf{T}^{W}_{a,k-1}(t)
\mathbf{T}^{\mathrm{orig}}_{a,k}
\mathbf{T}^{\mathrm{mot}}_{a,k}(q^{a}_{t,k}),
\]
where $\mathbf{T}^{\mathrm{orig}}_{a,k}$ is the static joint transform parsed from the URDF, and $\mathbf{T}^{\mathrm{mot}}_{a,k}$ is the revolute or prismatic motion induced by the joint value. For the gripper, the normalized signal is mapped to a prismatic displacement
\[
d^{a}_t=\mathrm{clip}(g^{a}_t,0,1)d^{a}_{\max},
\]
and used to compute the two finger branches. This gives a set of 3D robot keypoints
\[
\mathcal{P}^{a}_t=\{\mathbf{p}^{W}_{a,k}(t)\}_{k=1}^{K_a},
\]
including the arm skeleton, joint landmarks, and gripper endpoints.

\paragraph{Camera projection.}
Each 3D keypoint is projected to the target camera view. For a world-frame point $\mathbf{p}^{W}$, we compute
\[
\mathbf{p}^{C}
=
\mathbf{E}_t
\begin{bmatrix}
\mathbf{p}^{W}\\1
\end{bmatrix},
\qquad
\hat{\mathbf{u}}
=
\mathbf{K}_t\mathbf{p}^{C}_{1:3},
\]
and obtain the image coordinate by perspective division:
\[
(u,v)=
\left(
\frac{\hat{u}_x}{\hat{u}_z},
\frac{\hat{u}_y}{\hat{u}_z}
\right).
\]
Points with non-positive camera depth are discarded. This step maps robot kinematics from the world frame to the exact image plane used by the video world model.

\paragraph{Visual rendering.}
Given projected keypoints, we render a KVAF frame $\mathbf{V}_t\in\mathbb{R}^{H\times W\times 3}$ on a black canvas. The rendered elements include depth-aware arm skeletons, joint landmarks, gripper geometry, end-effector heatmaps, and pose axes. For depth-aware skeleton rendering, we normalize the camera depth $z$ within an episode-level range $(z_{\min},z_{\max})$ and map it to a color palette:
\[
\alpha(z)=
\mathrm{clip}
\left(
\frac{z-z_{\min}}{z_{\max}-z_{\min}},
0,1
\right),
\qquad
\mathbf{c}(z)=\mathcal{C}(\alpha(z)),
\]
where $\mathcal{C}$ denotes the depth color map. This allows the visual field to encode not only image-space position, but also relative depth ordering.

For the end-effector, we render a truncated Gaussian heatmap centered at the projected end-effector location $(u^a_t,v^a_t)$:
\[
H^{a}_t(x,y)
=
\exp
\left(
-\frac{(x-u^a_t)^2+(y-v^a_t)^2}{2\sigma^2}
\right)
\mathbf{1}
\left[
(x-u^a_t)^2+(y-v^a_t)^2\le r^2
\right].
\]
This heatmap provides an explicit cue for the manipulation focus. We further render the local end-effector pose axes. Given the rotation matrix $\mathbf{R}^{a}_t$ from the end-effector quaternion, each local axis endpoint is
\[
\mathbf{p}^{a,m}_t
=
\mathbf{p}^{a}_t
+
\ell_{\mathrm{axis}}\mathbf{R}^{a}_t\mathbf{e}_m,
\qquad
m\in\{x,y,z\}.
\]
After projection, the three axes are drawn in RGB colors. The final KVAF frame is the composition of all rendered components:
\[
\mathbf{V}_t
=
\mathrm{Render}
\left(
\mathcal{P}^{L}_t,\mathcal{P}^{R}_t,
\boldsymbol{\xi}^{L}_t,\boldsymbol{\xi}^{R}_t,
\mathbf{K}_t,\mathbf{E}_t
\right).
\]
Stacking all frames gives the KVAF sequence
\[
\mathcal{V}=\{\mathbf{V}_t\}_{t=1}^{T}.
\]
In the current implementation, KVAFs explicitly encode robot-side geometric motion information. Object-state changes are not directly rasterized in KVAFs, but are later captured through event-aware fusion and EDLS in the generative model.

\begin{algorithm}[t]
\caption{KVAF construction for one episode}
\label{alg:kvaf_construction}
\begin{algorithmic}[1]
\Require Robot URDF, embodiment config, episode states $\{\mathcal{I}_t\}_{t=1}^{T}$, image size $(H,W)$
\Ensure KVAF sequence $\mathcal{V}=\{\mathbf{V}_t\}_{t=1}^{T}$

\State Parse URDF to obtain joint origins, joint axes, link hierarchy, and gripper joints
\State Build left/right arm kinematic chains from the robot configuration
\State Estimate an episode-level depth range $(z_{\min},z_{\max})$ from visible robot keypoints
\For{$t=1$ to $T$}
    \State Initialize a black canvas $\mathbf{V}_t\in\mathbb{R}^{H\times W\times 3}$
    \For{each arm $a\in\{L,R\}$}
        \State Compute 3D arm and gripper keypoints by forward kinematics:
        \[
        \mathcal{P}^{a}_t
        \gets
        \mathrm{FK}
        (\mathbf{q}^{a}_t,g^{a}_t,\mathrm{URDF})
        \]
        \State Project keypoints to the target camera:
        \[
        \Pi^{a}_t
        \gets
        \mathrm{Project}
        (\mathcal{P}^{a}_t,\mathbf{K}_t,\mathbf{E}_t)
        \]
        \State Render depth-aware arm skeleton and joint landmarks on $\mathbf{V}_t$
        \State Render gripper geometry using projected finger keypoints
        \State Project end-effector center and render Gaussian heatmap
        \State Project end-effector pose axes and render RGB orientation cues
    \EndFor
    \State Append $\mathbf{V}_t$ to $\mathcal{V}$
\EndFor
\State \Return $\mathcal{V}$
\end{algorithmic}
\end{algorithm}

\begin{table}[H]
\centering
\caption{Rendered components in KVAFs and their roles.}
\label{tab:kvaf_components}
\small
\setlength{\tabcolsep}{5pt}
\begin{tabular}{ll}
\toprule
Component & Encoded information \\
\midrule
Depth-aware skeleton & Arm topology, spatial position, and relative depth \\
Joint landmarks & Local joint locations and articulated structure \\
Gripper geometry & End-effector opening state and finger configuration \\
End-effector heatmap & Manipulation focus and target interaction region \\
Pose axes & End-effector orientation and approach direction \\
\bottomrule
\end{tabular}
\end{table}

\subsection{Heuristic Action Recovery from KVAFs}
\label{app:kvaf_action_recovery}

This section describes the heuristic procedure used in our additional analysis to recover numerical actions from generated KVAF videos. This procedure is not part of EA-WM training or inference; it is only used to examine whether the visual action fields contain recoverable low-dimensional action information. Given a generated KVAF video, camera intrinsics and extrinsics, and the known rendered pose-axis length, we recover end-effector trajectories, gripper states, and then convert them into RoboTwin-style relative actions.

\paragraph{End-effector detection and pose recovery.}
For each KVAF frame, we first detect candidate end-effector centers from high-intensity heatmap regions. Around each detected center, we identify the rendered RGB pose-axis endpoints by thresholding local red, green, and blue regions. Let the detected image points be
\[
\mathbf{u}_{0},\mathbf{u}_{x},\mathbf{u}_{y},\mathbf{u}_{z}\in\mathbb{R}^{2},
\]
where $\mathbf{u}_{0}$ is the end-effector center and $\mathbf{u}_{x},\mathbf{u}_{y},\mathbf{u}_{z}$ correspond to the projected local $x,y,z$ axes. We associate them with the canonical 3D local points
\[
\mathbf{o}_{0}=\mathbf{0},\qquad
\mathbf{o}_{x}=\ell_{\mathrm{axis}}\mathbf{e}_{x},\qquad
\mathbf{o}_{y}=\ell_{\mathrm{axis}}\mathbf{e}_{y},\qquad
\mathbf{o}_{z}=\ell_{\mathrm{axis}}\mathbf{e}_{z}.
\]
Using the camera intrinsic matrix, we solve a PnP problem to estimate the end-effector pose in the camera frame, and then transform it to the world frame using the camera extrinsics. This gives a recovered pose
\[
\hat{\boldsymbol{\xi}}^{a}_{t}
=
(\hat{\mathbf{p}}^{a}_{t},\hat{\mathbf{R}}^{a}_{t})
\]
for each detected arm $a$ at time $t$.

\paragraph{Gripper state and bimanual track association.}
The gripper state is estimated from the local gripper geometry around the recovered end-effector center. We detect gripper-like white components near the pose axes and use their relative separation as a raw gripper opening signal. Since RoboTwin contains bimanual manipulation, detections are associated into left and right tracks. When two detections are present, we assign them by their recovered world-frame horizontal position. When only one detection is available, we assign it to the closer existing track. Missing detections are filled by nearest-neighbor interpolation, and the recovered pose and gripper sequences are smoothed with a short temporal median filter. Finally, the gripper curve is normalized to $[0,1]$ within each episode.

\paragraph{Conversion to relative actions.}
After obtaining dense left and right end-effector pose sequences, we convert them into RoboTwin-style relative actions. For each arm $a\in\{L,R\}$, the local translation action is computed in the current end-effector frame:
\[
\Delta \mathbf{p}^{a}_{t}
=
(\hat{\mathbf{R}}^{a}_{t})^{\top}
\left(
\hat{\mathbf{p}}^{a}_{t+1}-\hat{\mathbf{p}}^{a}_{t}
\right).
\]
The relative rotation is
\[
\Delta \mathbf{R}^{a}_{t}
=
(\hat{\mathbf{R}}^{a}_{t})^{\top}
\hat{\mathbf{R}}^{a}_{t+1},
\]
which is converted to roll-pitch-yaw angles:
\[
\Delta \boldsymbol{\theta}^{a}_{t}
=
\mathrm{RPY}(\Delta \mathbf{R}^{a}_{t}).
\]
The gripper action is the temporal difference of the normalized gripper state:
\[
\Delta g^{a}_{t}
=
\hat{g}^{a}_{t+1}-\hat{g}^{a}_{t}.
\]
Thus the recovered bimanual action at time $t$ is
\[
\hat{\mathbf{a}}_{t}
=
[
\Delta \mathbf{p}^{L}_{t},
\Delta \boldsymbol{\theta}^{L}_{t},
\Delta g^{L}_{t},
\Delta \mathbf{p}^{R}_{t},
\Delta \boldsymbol{\theta}^{R}_{t},
\Delta g^{R}_{t}
].
\]

\paragraph{Evaluation.}
We compare the recovered actions with ground-truth action CSVs using group-wise errors for translation, rotation, and gripper commands. Translation and rotation errors are computed as vector norms over the corresponding action dimensions, while the gripper error is computed as absolute error. Because the recovery depends on detecting small rendered cues such as heatmap centers, pose axes, and gripper components, the current heuristic pipeline is not expected to match direct numerical action prediction. Instead, it serves as an analysis tool for probing how much action information can be recovered from the image-domain KVAF representation.

\begin{algorithm}[t]
\caption{Heuristic action recovery from KVAF videos}
\label{alg:kvaf_action_recovery}
\begin{algorithmic}[1]
\Require Generated KVAF video $\mathcal{V}$, camera intrinsics $\mathbf{K}$, camera extrinsics $\mathbf{E}$, pose-axis length $\ell_{\mathrm{axis}}$
\Ensure Recovered relative action sequence $\{\hat{\mathbf{a}}_t\}$

\For{each frame $\mathbf{V}_t$}
    \State Detect candidate end-effector centers from heatmap peaks
    \For{each detected center}
        \State Detect local RGB pose-axis endpoints $(\mathbf{u}_x,\mathbf{u}_y,\mathbf{u}_z)$
        \State Solve PnP using canonical 3D axis points and detected 2D endpoints
        \State Convert the recovered camera-frame pose to world frame
        \State Estimate raw gripper opening from local gripper geometry
    \EndFor
    \State Store all valid detections for frame $t$
\EndFor

\State Associate detections into left/right tracks using recovered world positions
\State Fill missing detections by nearest-neighbor interpolation
\State Smooth pose and gripper trajectories with a temporal median filter
\State Normalize gripper curves to $[0,1]$

\For{each arm $a\in\{L,R\}$ and timestep $t$}
    \State Compute local translation:
    \[
    \Delta \mathbf{p}^{a}_{t}
    =
    (\hat{\mathbf{R}}^{a}_{t})^\top
    (\hat{\mathbf{p}}^{a}_{t+1}-\hat{\mathbf{p}}^{a}_{t})
    \]
    \State Compute local rotation:
    \[
    \Delta \boldsymbol{\theta}^{a}_{t}
    =
    \mathrm{RPY}
    \left(
    (\hat{\mathbf{R}}^{a}_{t})^\top
    \hat{\mathbf{R}}^{a}_{t+1}
    \right)
    \]
    \State Compute gripper delta:
    \[
    \Delta g^{a}_{t}
    =
    \hat{g}^{a}_{t+1}-\hat{g}^{a}_{t}
    \]
\EndFor

\State Concatenate left/right translation, rotation, and gripper deltas:
\[
\hat{\mathbf{a}}_{t}
=
[
\Delta \mathbf{p}^{L}_{t},
\Delta \boldsymbol{\theta}^{L}_{t},
\Delta g^{L}_{t},
\Delta \mathbf{p}^{R}_{t},
\Delta \boldsymbol{\theta}^{R}_{t},
\Delta g^{R}_{t}
]
\]
\State \Return $\{\hat{\mathbf{a}}_t\}$
\end{algorithmic}
\end{algorithm}

The relatively low detection rate of this heuristic recovery pipeline reflects the difficulty of inverting image-domain action fields back into precise numerical actions. This supports our view that KVAFs are better suited as a video-domain conditioning interface rather than as a direct replacement for numerical action labels.

\subsection{Full-metric evaluation.}
In the main paper, we focus on Physics Adherence, 3D Accuracy, and Controllability, as these dimensions are most directly related to robot motion, spatial consistency, and action-conditioned generation. For completeness, we additionally report all 16 WorldArena video perception metrics in Tables~\ref{tab:app_full_main_models} and~\ref{tab:app_full_action_models}. Table~\ref{tab:app_full_main_models} compares EA-WM with the representative video and embodied world models used in the main quantitative analysis, while Table~\ref{tab:app_full_action_models} compares KVAF-conditioned generation with mainstream action- or trajectory-conditioned baselines. Unlike the main-paper $\mathrm{P3CScore}$, which is computed only on the six selected metrics, the appendix $\mathrm{Avg}_{16}$ averages all 16 metrics including Action Following. Across both comparisons, EA-WM and KVAF-conditioned generation achieve the strongest overall $\mathrm{Avg}_{16}$, while also showing consistent advantages on robot-centric metrics such as Interaction Quality, Trajectory Accuracy, Depth Accuracy, and Perspectivity. These results further support that representing actions as camera-aligned KVAFs provides effective structured guidance for robotic video world modeling.

\begin{table}[H]
\centering
\caption{
Full WorldArena video perception metrics for the models compared in the main results.
All metrics are normalized to $[0,1]$ and higher is better.
$\mathrm{Avg}_{16}$ averages all 16 metrics, and multiplies the result by 100.
}
\label{tab:app_full_main_models}
\scriptsize
\setlength{\tabcolsep}{1.8pt}
\resizebox{\textwidth}{!}{
\begin{tabular}{lcccccccccccccccccc}
\toprule
& \multicolumn{3}{c}{Visual Quality}
& \multicolumn{3}{c}{Motion Quality}
& \multicolumn{3}{c}{Content Consistency}
& \multicolumn{2}{c}{Physics}
& \multicolumn{2}{c}{3D}
& \multicolumn{3}{c}{Controllability}
& \multirow{2}{*}{$\mathrm{Avg}_{16}$} \\
\cmidrule(lr){2-4}
\cmidrule(lr){5-7}
\cmidrule(lr){8-10}
\cmidrule(lr){11-12}
\cmidrule(lr){13-14}
\cmidrule(lr){15-17}
Model
& Img. & Aes. & JEPA
& Dyn. & Flow & Smooth
& Subj. & Bg. & Photo
& Inter. & Traj.
& Depth & Persp.
& Instr. & Sem. & Action
&  \\
\midrule
CogVideoX
& 0.358 & 0.378 & 0.938
& 0.317 & 0.219 & 0.739
& 0.808 & 0.877 & 0.358
& 0.594 & 0.353
& 0.910 & 0.783
& 0.727 & \textbf{0.898} & 0.008
& 57.91 \\

WoW
& 0.459 & 0.387 & 0.744
& 0.461 & 0.271 & 0.769
& 0.816 & 0.903 & 0.217
& 0.556 & 0.206
& 0.728 & 0.767
& 0.569 & 0.884 & 0.043
& 54.87 \\

Wan 2.2
& 0.388 & 0.396 & 0.758
& 0.435 & 0.127 & 0.702
& \textbf{0.839} & 0.904 & \textbf{0.478}
& 0.518 & 0.163
& 0.777 & 0.766
& 0.538 & 0.888 & 0.051
& 54.55 \\

TesserAct
& 0.332 & \textbf{0.459} & 0.458
& 0.515 & 0.245 & 0.758
& 0.825 & \textbf{0.924} & 0.249
& 0.580 & 0.140
& 0.716 & 0.792
& 0.615 & 0.878 & 0.031
& 53.23 \\

GigaWorld-0
& 0.504 & 0.399 & 0.441
& 0.671 & 0.312 & 0.781
& 0.730 & 0.856 & 0.176
& 0.537 & 0.155
& 0.632 & 0.760
& 0.616 & 0.859 & 0.113
& 53.39 \\

Vidar
& 0.415 & 0.407 & 0.561
& 0.277 & 0.143 & \textbf{0.797}
& 0.763 & 0.830 & 0.235
& 0.535 & 0.193
& 0.787 & 0.759
& 0.591 & 0.883 & 0.082
& 51.61 \\

Cosmos-Predict 2.5 (text)
& \textbf{0.667} & 0.450 & 0.313
& 0.591 & \textbf{0.430} & 0.788
& 0.749 & 0.851 & 0.138
& 0.387 & 0.082
& 0.705 & 0.796
& 0.266 & 0.773 & \textbf{0.142}
& 50.80 \\

Genie Envisioner
& 0.231 & 0.329 & 0.334
& \textbf{0.693} & 0.086 & 0.697
& 0.776 & 0.902 & 0.201
& 0.205 & 0.068
& 0.866 & 0.528
& 0.203 & 0.854 & 0.011
& 43.65 \\

\midrule
\textbf{EA-WM (Ours)}
& 0.364 & 0.387 & \textbf{0.951}
& 0.360 & 0.273 & 0.730
& 0.827 & 0.904 & 0.220
& \textbf{0.682} & \textbf{0.430}
& \textbf{0.959} & \textbf{0.838}
& \textbf{0.792} & 0.895 & 0.048
& \textbf{60.38} \\
\bottomrule
\end{tabular}
}
\end{table}

\begin{table}[H]
\centering
\caption{
Full WorldArena video perception metrics for mainstream action- or trajectory-conditioned baselines.
All metrics are normalized to $[0,1]$ and higher is better.
$\mathrm{Avg}_{16}$ averages all 16 metrics, and multiplies the result by 100.
}
\label{tab:app_full_action_models}
\scriptsize
\setlength{\tabcolsep}{1.8pt}
\resizebox{\textwidth}{!}{
\begin{tabular}{lcccccccccccccccccc}
\toprule
& \multicolumn{3}{c}{Visual Quality}
& \multicolumn{3}{c}{Motion Quality}
& \multicolumn{3}{c}{Content Consistency}
& \multicolumn{2}{c}{Physics}
& \multicolumn{2}{c}{3D}
& \multicolumn{3}{c}{Controllability}
& \multirow{2}{*}{$\mathrm{Avg}_{16}$} \\
\cmidrule(lr){2-4}
\cmidrule(lr){5-7}
\cmidrule(lr){8-10}
\cmidrule(lr){11-12}
\cmidrule(lr){13-14}
\cmidrule(lr){15-17}
Model
& Img. & Aes. & JEPA
& Dyn. & Flow & Smooth
& Subj. & Bg. & Photo
& Inter. & Traj.
& Depth & Persp.
& Instr. & Sem. & Action
&  \\
\midrule
CtrlWorld
& 0.352 & \textbf{0.389} & 0.919
& 0.426 & \textbf{0.345} & \textbf{0.738}
& \textbf{0.841} & 0.906 & 0.173
& 0.621 & 0.477
& 0.930 & 0.796
& 0.727 & 0.891 & 0.021
& 59.70 \\

IRASim
& 0.349 & 0.362 & 0.933
& 0.414 & 0.208 & 0.705
& 0.831 & 0.907 & 0.352
& 0.566 & 0.364
& 0.931 & 0.779
& 0.660 & 0.885 & \textbf{0.053}
& 58.12 \\

Cosmos-Predict 2.5 (action)
& \textbf{0.449} & 0.358 & 0.930
& 0.399 & 0.057 & 0.710
& 0.820 & 0.889 & \textbf{0.353}
& 0.550 & 0.295
& 0.886 & 0.764
& 0.584 & 0.888 & 0.013
& 55.91 \\

RoboMaster
& 0.349 & 0.384 & 0.297
& \textbf{0.612} & 0.148 & 0.694
& 0.830 & \textbf{0.912} & 0.336
& 0.536 & 0.116
& 0.834 & 0.759
& 0.577 & 0.876 & 0.035
& 51.84 \\

\midrule
\textbf{KVAF-conditioned generation}
& 0.364 & 0.383 & \textbf{0.950}
& 0.360 & 0.268 & 0.729
& 0.818 & 0.899 & 0.229
& \textbf{0.682} & \textbf{0.494}
& \textbf{0.977} & \textbf{0.850}
& \textbf{0.784} & \textbf{0.901} & 0.048
& \textbf{60.84} \\
\bottomrule
\end{tabular}
}
\end{table}

\subsection{Additional Visualizations of Generated KVAFs and Videos}
\label{app:kvaf_video_visualization}

\begin{figure}[!t]
\centering
\includegraphics[width=\textwidth]{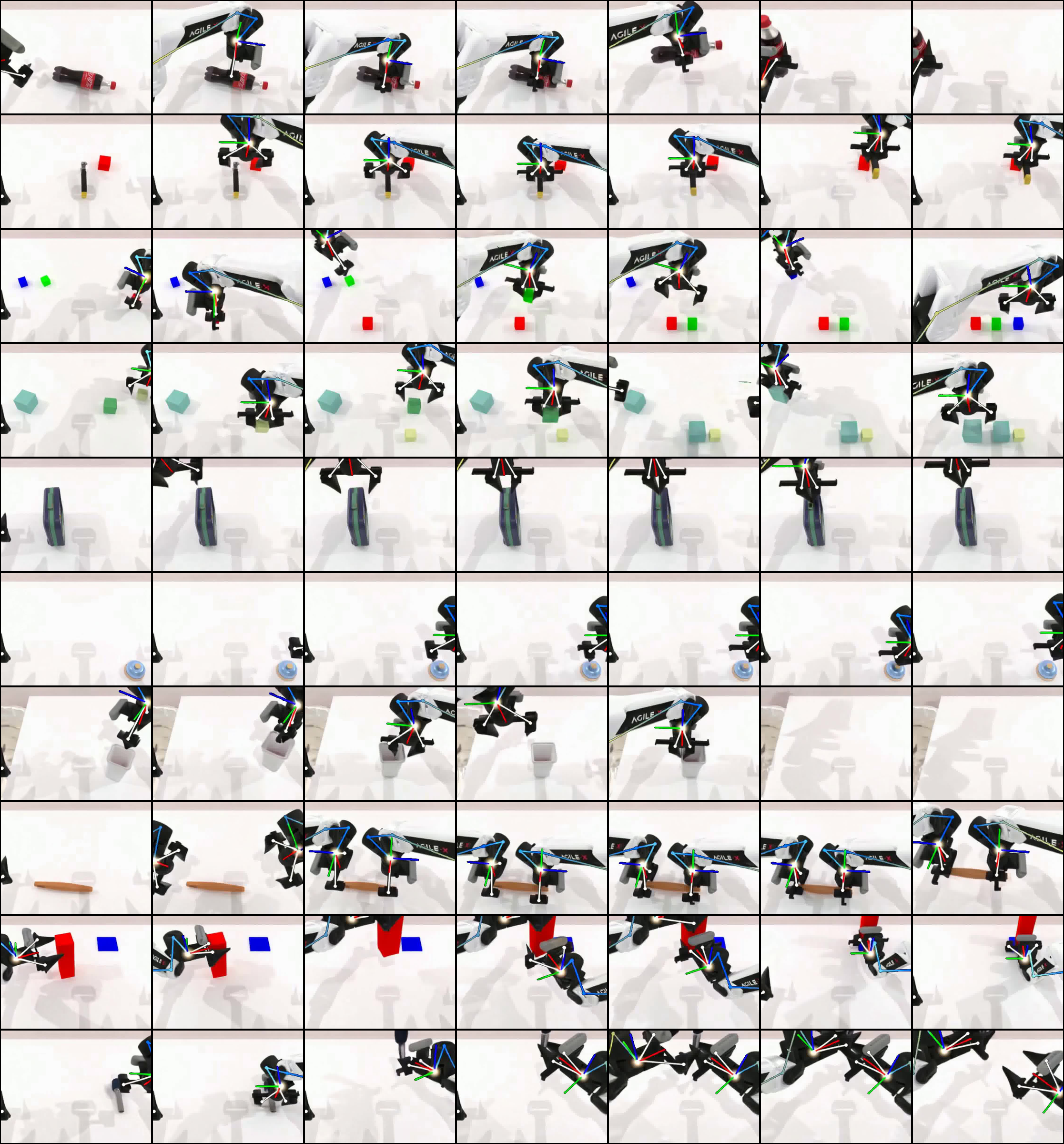}
\caption{
Generated KVAF and video frame sequence
}
\label{fig:app_kvaf_video_6}
\end{figure}

\begin{figure}[!t]
\centering
\includegraphics[width=\textwidth]{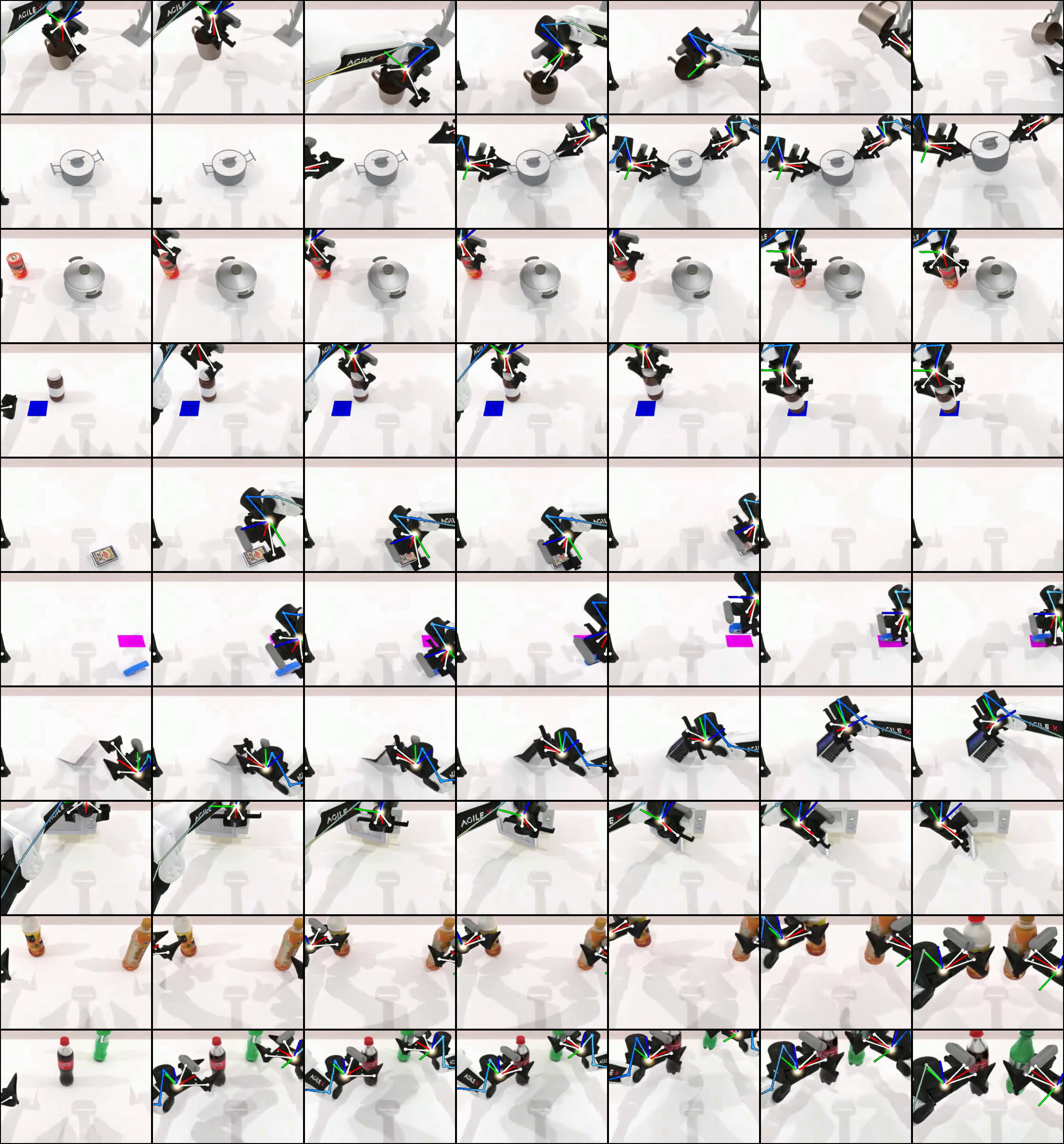}
\caption{
Generated KVAF and video frame sequence
}
\label{fig:app_kvaf_video_7}
\end{figure}

\begin{figure}[!t]
\centering
\includegraphics[width=\textwidth]{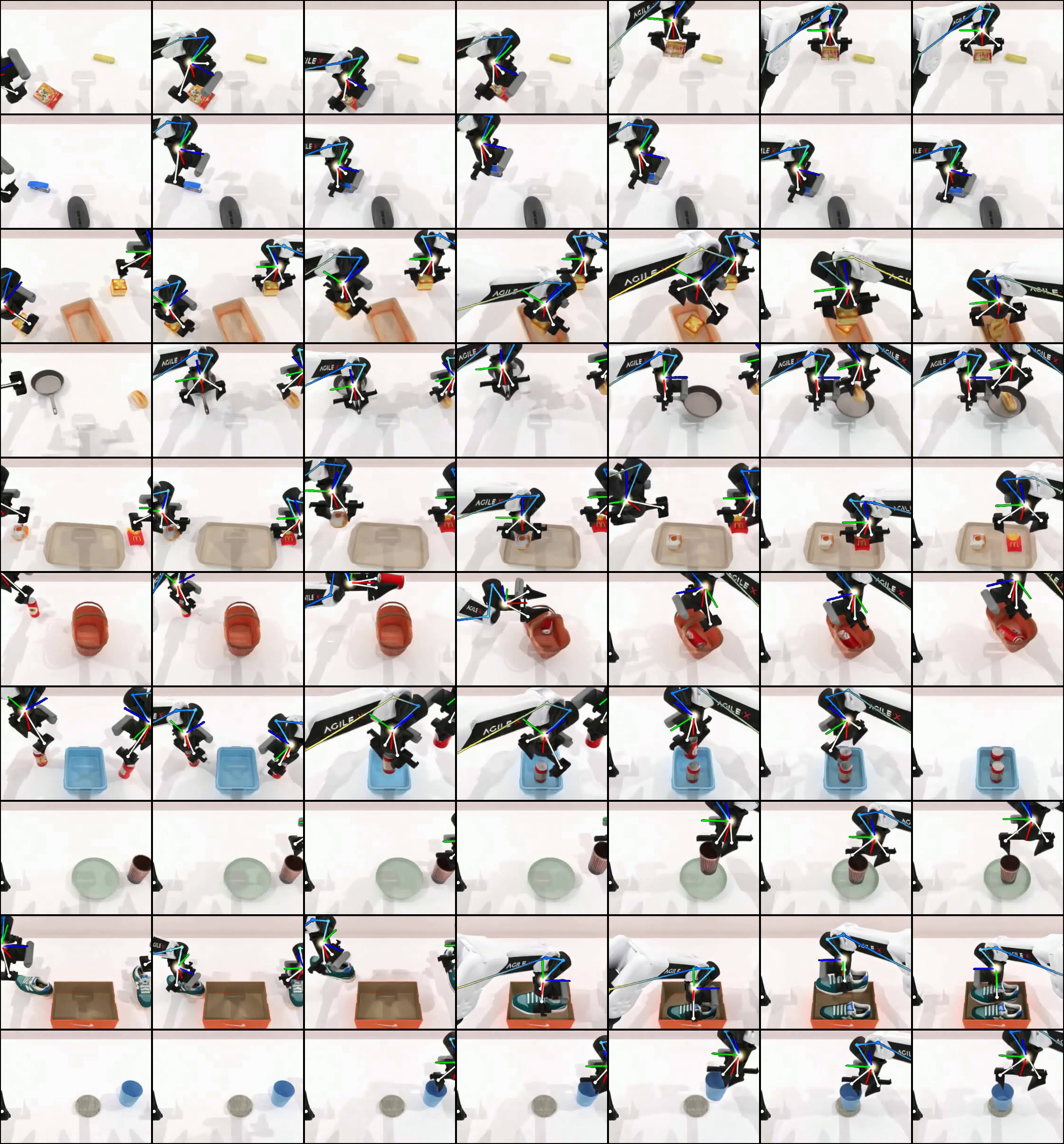}
\caption{
Generated KVAF and video frame sequence
}
\label{fig:app_kvaf_video_8}
\end{figure}

\begin{figure}[!t]
\centering
\includegraphics[width=\textwidth]{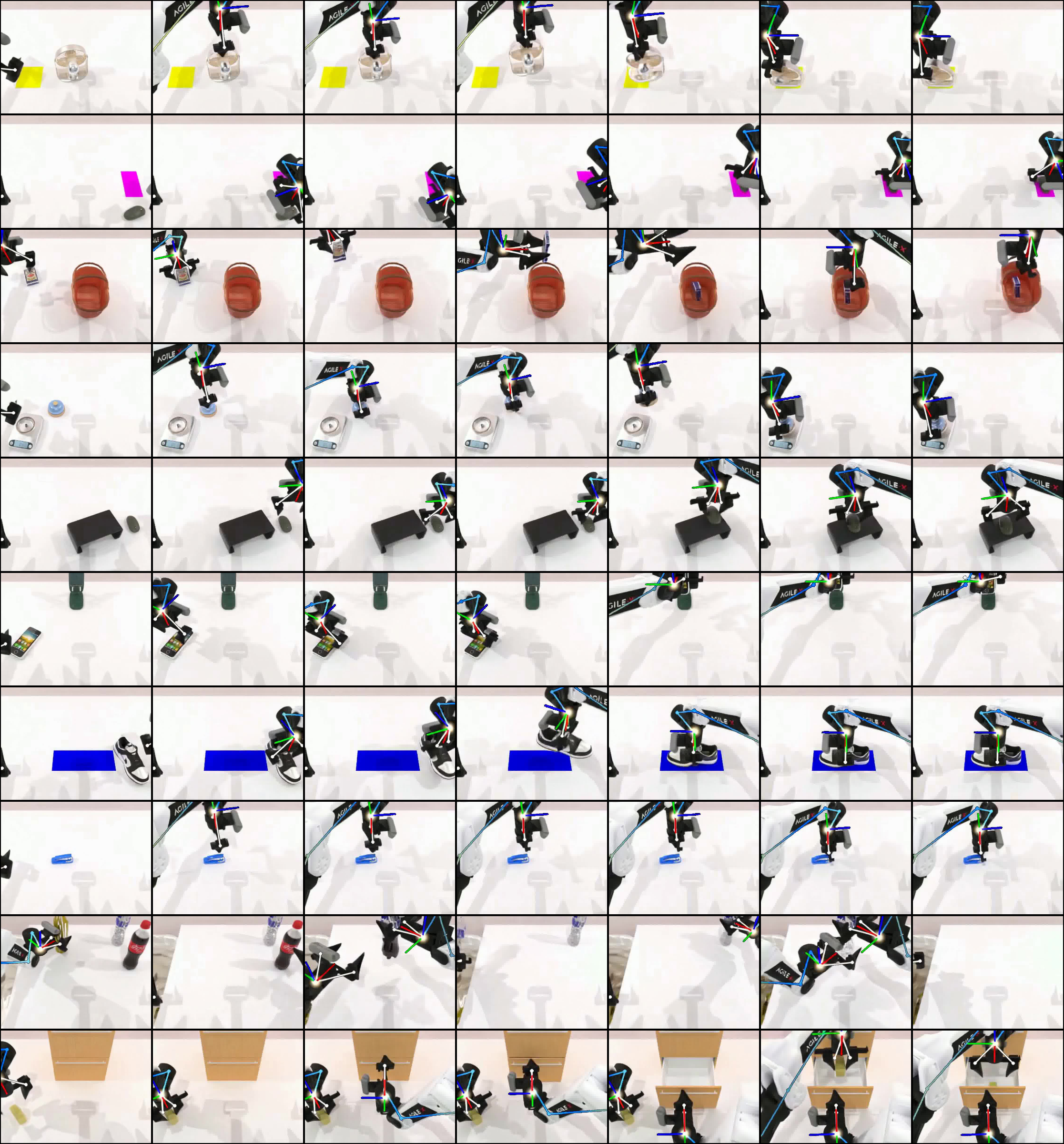}
\caption{
Generated KVAF and video frame sequence
}
\label{fig:app_kvaf_video_9}
\end{figure}

\begin{figure}[!t]
\centering
\includegraphics[width=\textwidth]{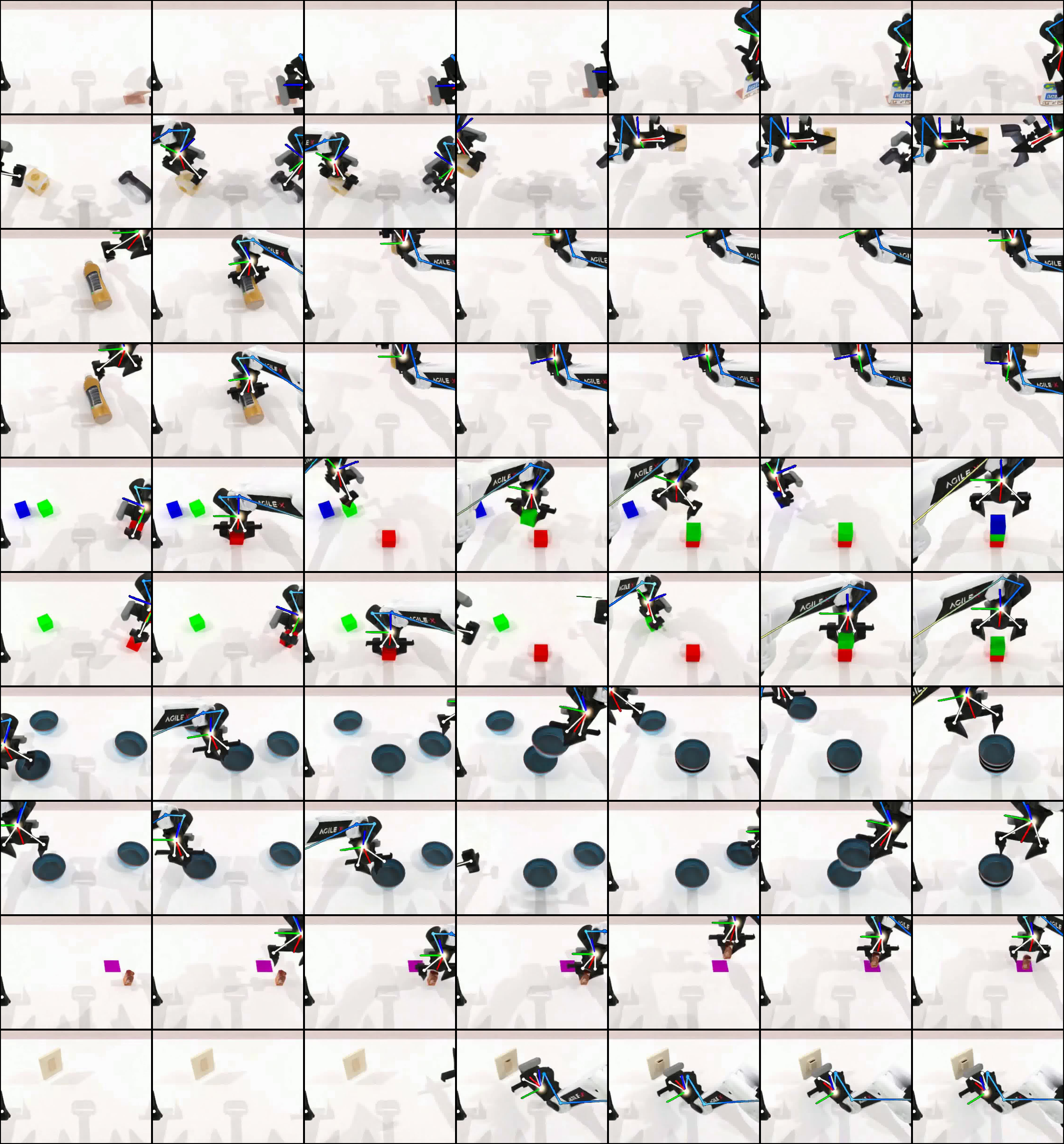}
\caption{
Generated KVAF and video frame sequence
}
\label{fig:app_kvaf_video_10}
\end{figure}

To further examine the learned video-domain action representation, we visualize generated KVAF sequences together with the corresponding generated video frames in Figures~\ref{fig:app_kvaf_video_6}--\ref{fig:app_kvaf_video_10}. Across different tasks, the generated KVAFs are well aligned with the robot motion in the videos: skeletons, end-effector cues, gripper geometry, and pose axes follow the generated robot configurations, while the video frames maintain consistent object locations and interaction progress.

\subsection{Broader Impacts}
\label{app:broader_impacts}

EA-WM is primarily evaluated in simulation, with a preliminary offline evaluation on real-world RH20T data. The model is not deployed for closed-loop control on physical robots in this work. The model does not directly execute robot actions, nor is it deployed on real robots in this work. Nevertheless, more accurate robotic world models may eventually be used as predictive simulators, planning modules, policy-learning aids, or evaluation environments for downstream VLA systems. If such models generate misleading future rollouts, downstream policies or planners may make unsafe decisions, especially in physical environments involving humans, fragile objects, or safety-critical manipulation.

This work does not involve human-subject data, personally identifiable information, or real-robot deployment. The primary goal is to study how structured action information can improve robotic video world modeling.

\subsection{Controlled Analysis of KVAF Representations}
\label{sec:kvaf_representation_analysis}

We further conduct controlled experiments to disentangle the effect of the KVAF representation from additional geometric information and model capacity. \textit{Geometry49} contains the same robot kinematic states, end-effector poses, gripper states, and camera parameters used to construct KVAFs, but encodes them as a 49-dimensional numerical representation. We additionally consider two simplified image-space representations: \textit{EEF Pose Map}, which retains only projected end-effector locations and orientations, and \textit{Binary Embodiment Mask}, which rasterizes the projected robot structure as a binary mask. The visual representation variants use the same dual-stream architecture, fusion design, training configuration, and parameter budget, differing only in the information retained by the visual action representation.

\begin{table}[!t]
\caption{
Full WorldArena comparison of KVAF with an information-matched numerical representation and simplified visual representations. All individual metrics are normalized to $[0,1]$ and higher is better. EWMScore averages all 16 WorldArena metrics and is multiplied by 100; P3CScore averages the six robot-centric metrics used in the main paper and is also multiplied by 100. Aggregate scores are computed from the unrounded metric values.
}
\label{tab:kvaf_representation_full}
\centering
\small
\setlength{\tabcolsep}{5.0pt}
\begin{tabular}{lcccc}
\toprule
Metric & Full KVAF & Geometry49 & Binary Mask & EEF Pose Map \\
\midrule
\multicolumn{5}{l}{\textit{Visual Quality}} \\
Image Quality              & 0.364 & 0.367 & 0.360 & \textbf{0.376} \\
Aesthetic Quality          & 0.387 & 0.383 & 0.383 & \textbf{0.389} \\
JEPA Similarity            & \textbf{0.951} & 0.830 & 0.939 & 0.932 \\
\midrule
\multicolumn{5}{l}{\textit{Motion Quality}} \\
Dynamic Degree             & 0.360 & \textbf{0.361} & 0.345 & 0.325 \\
Flow Score                 & \textbf{0.273} & 0.254 & 0.258 & 0.245 \\
Motion Smoothness          & 0.730 & \textbf{0.740} & 0.725 & 0.690 \\
\midrule
\multicolumn{5}{l}{\textit{Content Consistency}} \\
Subject Consistency        & \textbf{0.827} & 0.804 & 0.812 & 0.822 \\
Background Consistency     & \textbf{0.904} & 0.894 & 0.891 & 0.893 \\
Photometric Consistency    & 0.220 & 0.231 & \textbf{0.256} & 0.248 \\
\midrule
\multicolumn{5}{l}{\textit{Physics Adherence}} \\
Interaction Quality        & 0.682 & 0.666 & 0.682 & \textbf{0.696} \\
Trajectory Accuracy        & \textbf{0.430} & 0.230 & 0.357 & 0.282 \\
\midrule
\multicolumn{5}{l}{\textit{3D Accuracy}} \\
Depth Accuracy             & \textbf{0.959} & 0.847 & 0.932 & 0.892 \\
Perspectivity              & \textbf{0.838} & 0.828 & 0.836 & 0.812 \\
\midrule
\multicolumn{5}{l}{\textit{Controllability}} \\
Instruction Following      & \textbf{0.792} & 0.758 & 0.780 & 0.716 \\
Semantic Alignment         & \textbf{0.895} & 0.878 & 0.888 & 0.880 \\
Action Following & 0.0480 & 0.0476 & 0.0475 & \textbf{0.0491} \\
\midrule
\textbf{EWMScore}          & \textbf{60.38} & 56.99 & 59.32 & 57.79 \\
\textbf{P3CScore}          & \textbf{76.60} & 70.11 & 74.60 & 71.30 \\
\bottomrule
\end{tabular}

\vspace{2pt}
\end{table}

Despite containing the same underlying robot and camera information, Geometry49 is substantially weaker than Full KVAF, most notably on Trajectory Accuracy (0.230 vs.\ 0.430) and Depth Accuracy (0.847 vs.\ 0.959), resulting in a 6.49-point P3CScore gap. The improvement is also concentrated in robot-centric geometry rather than generic visual quality: compared with Geometry49, Binary Mask changes Image Quality, Aesthetic Quality, and Background Consistency only marginally, while improving Trajectory Accuracy by 12.72 points and Depth Accuracy by 8.54 points. This suggests that camera-aligned spatial organization, rather than privileged geometric information alone, is critical for making robot motion and 3D structure easier for the video model to exploit.

\clearpage
\begin{figure}[H]
    \centering
    \includegraphics[width=\linewidth]{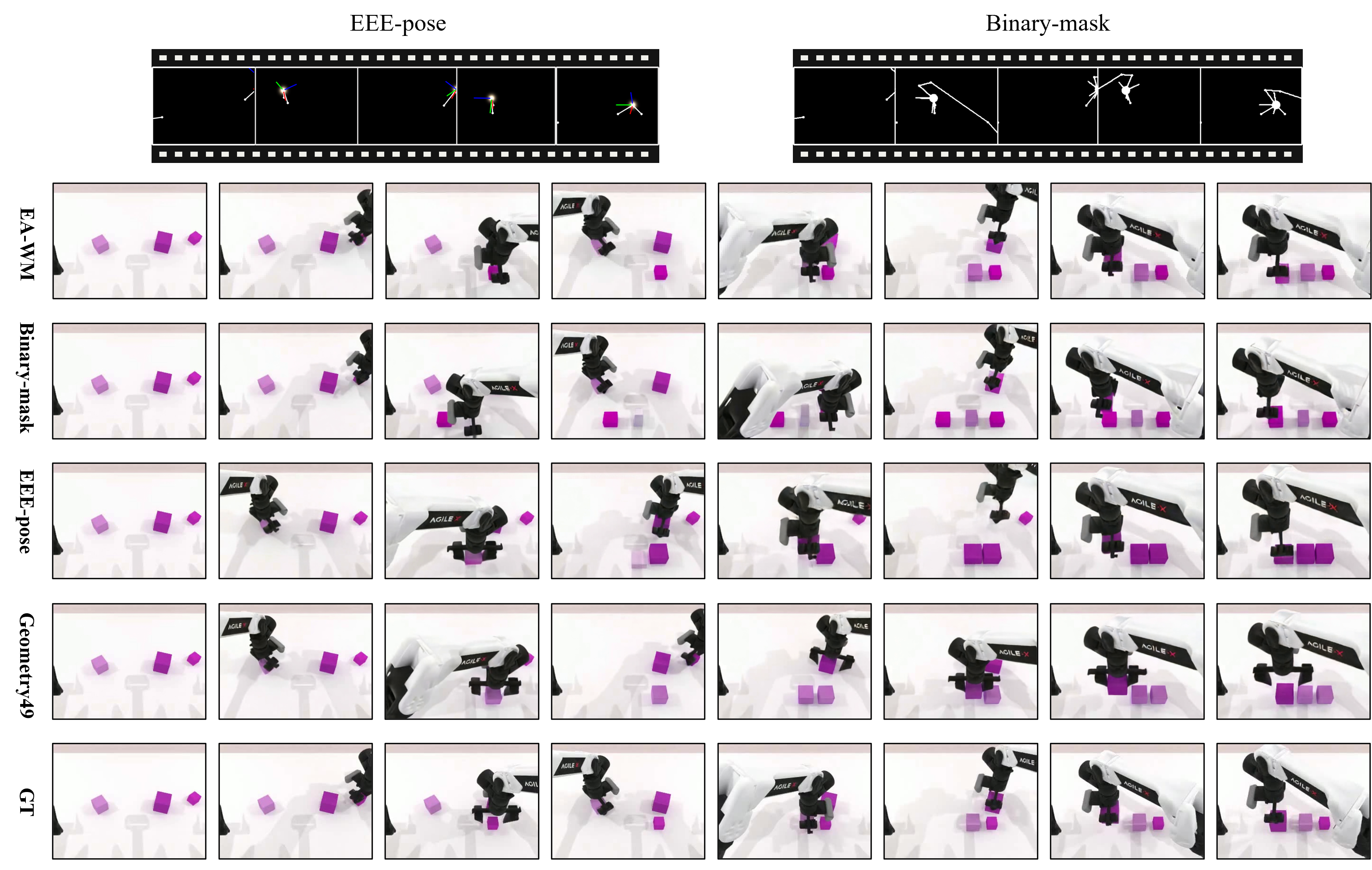}
    \caption{
    Qualitative comparison of simplified visual action representations and their corresponding RGB rollouts. EEF Pose Map provides sparse end-effector position and orientation cues, whereas Binary Embodiment Mask additionally exposes coarse full-arm geometry in the camera view.
    }
    \label{fig:representation_controls}
\end{figure}

The simplified visual controls show a consistent hierarchy. EEF Pose Map remains competitive on local Interaction Quality (0.696), but its sparse endpoint cues leave the global arm trajectory and depth structure under-constrained, yielding 0.282 Trajectory Accuracy and 0.892 Depth Accuracy. Binary Mask improves these metrics to 0.357 and 0.932 by explicitly constraining coarse full-arm occupancy, consistent with the qualitative rollouts in Fig.~\ref{fig:representation_controls}. Full KVAF further reaches 0.430 and 0.959 by retaining articulated topology, relative depth, joint locations, gripper state, manipulation focus, and end-effector pose. These results indicate that the advantage of KVAF comes from the structured kinematic content retained in the representation, rather than simply from additional model capacity or rasterizing robot motion into image space.

\subsection{Evaluation Protocol and Baseline Reproduction}
\label{sec:evaluation_protocol}

All EA-WM results, ablations, and locally reproduced baselines are evaluated on the released WorldArena validation set using the same evaluation pipeline and preprocessing. For external baselines, we use the results reported by WorldArena, which evaluates representative video and embodied world models under a unified benchmark protocol. We retain these reported values to maintain consistent provenance across the comparison tables. Since complete model-specific post-training configurations are not available for every baseline, these comparisons should be interpreted as benchmark-level comparisons under a common data split and evaluator, rather than strictly parameter- or optimization-matched comparisons.

As an additional sanity check, we locally post-trained both Wan2.2 and CtrlWorld on the released RoboTwin split using the same overall data and optimization budget as our main experiments, while retaining their original architectures and conditioning interfaces. The reproduced results closely match the corresponding WorldArena values, as summarized in Table~\ref{tab:baseline_reproduction}. The small discrepancies observed for both a text-conditioned model and an action-conditioned model suggest that our local evaluation setup is well aligned with the benchmark protocol. We therefore use the WorldArena-reported numbers for external baselines in the main comparisons and use local reproductions primarily as protocol verification.

\begin{table}[H]
\caption{
Local verification of the WorldArena evaluation protocol. Values report EWMScore; the reproduced Wan2.2 and CtrlWorld results closely match the corresponding WorldArena values.
}
\label{tab:baseline_reproduction}
\centering
\small
\setlength{\tabcolsep}{7pt}
\begin{tabular}{lccc}
\toprule
Model & WorldArena & Local reproduction & Abs. difference \\
\midrule
Wan2.2    & 54.55 & 54.62 & 0.07 \\
CtrlWorld & 59.70 & 58.97 & 0.73 \\
\bottomrule
\end{tabular}
\end{table}

EA-WM is evaluated under two separately trained configurations, depending on whether future KVAFs are generated jointly with RGB video or supplied as action-derived conditions. In the main joint-generation setting, the model receives only the initial frame and text instruction; both future RGB and KVAF latents are initialized from noise and generated jointly. This setting is compared with text-conditioned world models. We additionally train a KVAF-conditioned variant for comparison with action- or trajectory-conditioned models. In this setting, the action trajectory is first converted into reference KVAFs, which are provided to the model as conditions for RGB generation, while the KVAF prediction head is removed.

\begin{table}[H]
\caption{
Two evaluation configurations of EA-WM. The joint-generation setting evaluates future RGB--KVAF co-generation from an initial observation and text instruction, whereas the KVAF-conditioned setting evaluates KVAFs as an explicit action-conditioning interface.
}
\label{tab:eawm_settings}
\centering
\small
\setlength{\tabcolsep}{3.5pt}
\begin{tabular}{
p{0.18\linewidth}
p{0.25\linewidth}
p{0.28\linewidth}
p{0.17\linewidth}}
\toprule
Setting & Inference input & KVAF role & Comparison \\
\midrule
Joint-generation EA-WM
& Initial frame + text
& Future RGB and KVAF latents are jointly generated from noise
& Text-conditioned models \\
\addlinespace
KVAF-conditioned EA-WM
& Initial frame + text + action-derived KVAFs
& Reference KVAFs condition RGB generation
& Action-/trajectory-conditioned models \\
\bottomrule
\end{tabular}
\end{table}

This distinction is important when interpreting the two comparison groups: the joint-generation setting evaluates whether EA-WM can jointly model future visual and kinematic evolution from an initial observation and language instruction, whereas the KVAF-conditioned setting evaluates the effectiveness of KVAFs as an explicit action-conditioning interface. The latter therefore has access to a more informative action-derived condition and is not intended as a directly matched replacement for the joint-generation setting. Together, the two evaluations characterize EA-WM both as a generative world model and as an action-conditioned video predictor.

\subsection{Computational Overhead}
\label{sec:computational_overhead}

EA-WM introduces a second full-depth DiT stream for modeling KVAFs, and therefore has a larger computational footprint than the original single-stream Wan2.2-TI2V-5B backbone. We quantify this overhead in terms of model size, inference latency, GPU memory usage, and KVAF construction cost.

\paragraph{Model size and inference overhead.}
We benchmark Wan2.2-TI2V-5B and EA-WM on a single NVIDIA A100-SXM4-80GB GPU using bfloat16, $480\times640$ resolution, and 50 denoising steps. All latency results are averaged over three runs after warm-up and include preprocessing, denoising, and RGB VAE decoding.

\begin{table}[H]
\caption{
Computational overhead of EA-WM relative to Wan2.2-TI2V-5B.
DiT-side parameter counts exclude the shared video VAE and text encoder.
}
\label{tab:computational_overhead}
\centering
\small
\setlength{\tabcolsep}{7pt}
\begin{tabular}{lccc}
\toprule
Method
& DiT-side Params.
& Latency (s)
& Peak Memory (GiB) \\
\midrule
Wan2.2-TI2V-5B
& 5.00B
& $49.30 \pm 0.08$
& 24.79 \\

EA-WM (RGB)
& 10.40B
& $96.38 \pm 0.05$
& 34.88 \\

EA-WM (RGB + KVAF decoding)
& 10.40B
& $100.89 \pm 0.07$
& 34.88 \\
\bottomrule
\end{tabular}
\end{table}

EA-WM contains 10.40B DiT-side parameters, compared with 5.00B for Wan2.2-TI2V-5B. Most of the additional parameters come from the full-depth KVAF branch. For RGB generation, inference latency increases from 49.30\,s to 96.38\,s, corresponding to an increase of 47.08\,s (95.5\%), while peak allocated GPU memory increases from 24.79\,GiB to 34.88\,GiB, an increase of 10.09\,GiB (40.7\%). We therefore do not claim computational or parameter parity with the single-stream Wan2.2 backbone. Decoding the generated KVAF visualization is optional and adds only 4.51\,s without further increasing peak GPU memory.

Importantly, the architectural overhead above should be distinguished from the contribution of the KVAF representation itself. In the capacity-controlled representation analysis in Sec.~\ref{sec:kvaf_representation_analysis}, we keep the dual-stream architecture, fusion design, training configuration, and parameter budget fixed while changing only the visual action representation. These controlled experiments show that the performance improvement cannot be explained solely by the increased model capacity.

\paragraph{KVAF construction cost.}
We separately benchmark KVAF construction on an Intel Xeon Platinum 8362 CPU. The measurement includes HDF5 loading, forward-kinematic projection, depth preprocessing, visual-field rendering, and video encoding. The measured processing times are 0.113\,s, 0.401\,s, 0.722\,s, and 0.879\,s for trajectories of 9, 33, 65, and 81 frames, respectively.

\begin{figure}[H]
    \centering
    \includegraphics[width=\linewidth]{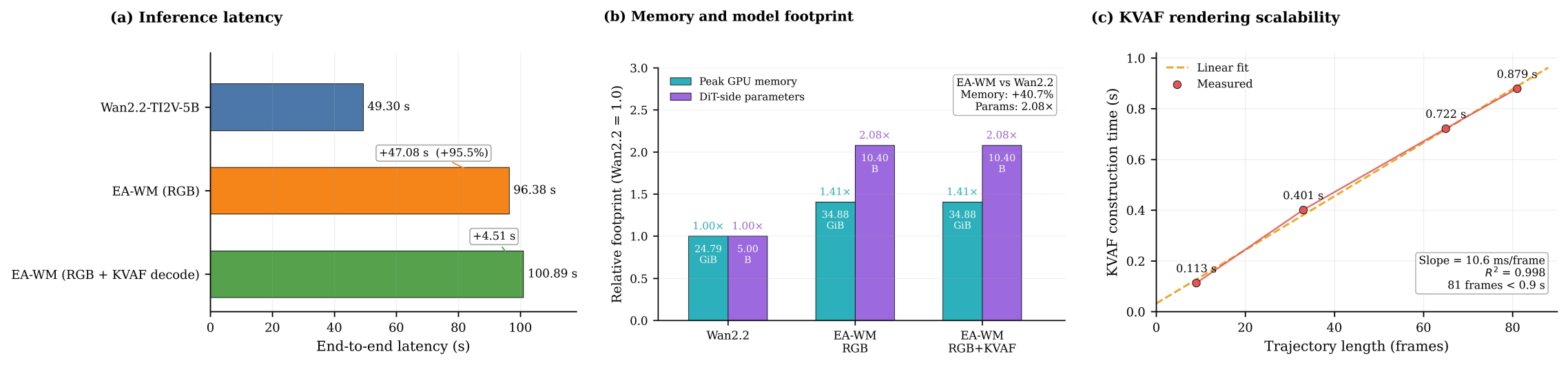}
    \caption{
    Computational profile of EA-WM.
    \textbf{(a)} End-to-end inference latency under the same sampling configuration. 
    EA-WM introduces a 95.5\% latency increase over Wan2.2-TI2V-5B, while optional KVAF decoding adds only 4.51\,s.
    \textbf{(b)} Peak allocated GPU memory and DiT-side parameter footprint. 
    Most of the additional computational footprint comes from the full-depth KVAF stream.
    \textbf{(c)} CPU cost of KVAF construction as trajectory length increases. 
    Rendering scales approximately linearly at 10.6\,ms per frame ($R^2=0.998$), with an 81-frame sequence requiring 0.879\,s.
    }
    \label{fig:computational_overhead}
\end{figure}

As shown in Fig.~\ref{fig:computational_overhead}(b), KVAF construction scales approximately linearly with trajectory length, at about 10.6\,ms per frame. Even an 81-frame trajectory requires less than 0.9\,s to construct, indicating that KVAF rendering itself accounts for only a small fraction of the end-to-end computational cost. The dominant overhead of the current implementation instead comes from iterative denoising through the additional full-depth DiT stream.

Moreover, in the main joint-generation setting, no reference future KVAF trajectory is required at inference: both future RGB and KVAF latents are initialized from noise and jointly generated. Therefore, reference KVAF construction does not introduce an additional online conditioning cost in the main EA-WM setting.

Overall, the current implementation incurs substantial overhead primarily because of the full-depth dual-stream architecture, rather than the construction of KVAFs themselves. At the same time, the capacity-controlled experiments in Sec.~\ref{sec:kvaf_representation_analysis} separate this architectural cost from the benefit of the structured KVAF representation. Reducing the computational footprint through a shallower or weight-shared KVAF branch, fewer fusion layers, or model distillation is an important direction for future work.

\subsection{Functional Utility of EA-WM}
\label{sec:functional_utility}

Beyond perceptual evaluation, we further examine whether the generated KVAF and RGB rollouts can support downstream robot learning, policy evaluation, and action planning.

\textbf{Data Engine.}
Following the WorldArena data-engine setting, we evaluate on
\texttt{adjust\_bottle} and \texttt{click\_bell}. We use a lightweight
temporal Transformer to decode generated KVAF sequences into robot actions.
Each 81-frame RGB KVAF sequence is encoded frame-wise by a small convolutional
encoder into 128-dimensional tokens, followed by a 4-layer, 4-head Transformer
encoder and an MLP head that predicts an 80-step, 14-dimensional action sequence.
The decoder is trained on ground-truth KVAF--action pairs and kept fixed for
downstream evaluation. The decoded trajectories are subsequently used to
fine-tune the $\pi_{0.5}$ policy.

\begin{table}[H]
\caption{
Data-engine evaluation on the two WorldArena tasks. Actions decoded from generated KVAF trajectories provide effective supervision for downstream policy learning.
}
\label{tab:data_engine}
\centering
\small
\setlength{\tabcolsep}{7pt}
\begin{tabular}{lccc}
\toprule
Training data & \texttt{adjust\_bottle} & \texttt{click\_bell} & Average \\
\midrule
Zero-shot $\pi_{0.5}$ & 2\% & 5\% & 3.5\% \\
WoW & 45\% & 71\% & 58.0\% \\
GT RoboTwin data & 77\% & 66\% & 71.5\% \\
Generated KVAF $\rightarrow$ actions & 75\% & 69\% & \textbf{72.0\%} \\
\bottomrule
\end{tabular}
\end{table}

Actions decoded from generated KVAFs achieve 72.0\% average success, compared with 58.0\% for the strongest reported world-model data engine and 71.5\% for the ground-truth RoboTwin-data reference on these two tasks. This result suggests that the jointly generated KVAF stream retains substantial executable information and can serve as a synthetic interface for downstream policy learning.

\textbf{Policy Evaluation.}
Following the WorldArena policy-evaluation protocol, we evaluate five
$\pi_{0.5}$ policies trained with 10\%, 20\%, 30\%, 50\%, and 100\% of the
training data. Each policy is evaluated over the full RoboTwin task set with
10 episodes per task, resulting in 500 rollouts per policy. We compare the
execution success measured in RoboTwin with the success estimated from EA-WM
rollouts using the WorldArena VLM-based success evaluator. The two measurements
exhibit strong correlation, with Pearson $r=0.922$ and Spearman $\rho=0.800$.

\begin{figure}[H]
    \centering
    \includegraphics[width=\linewidth]{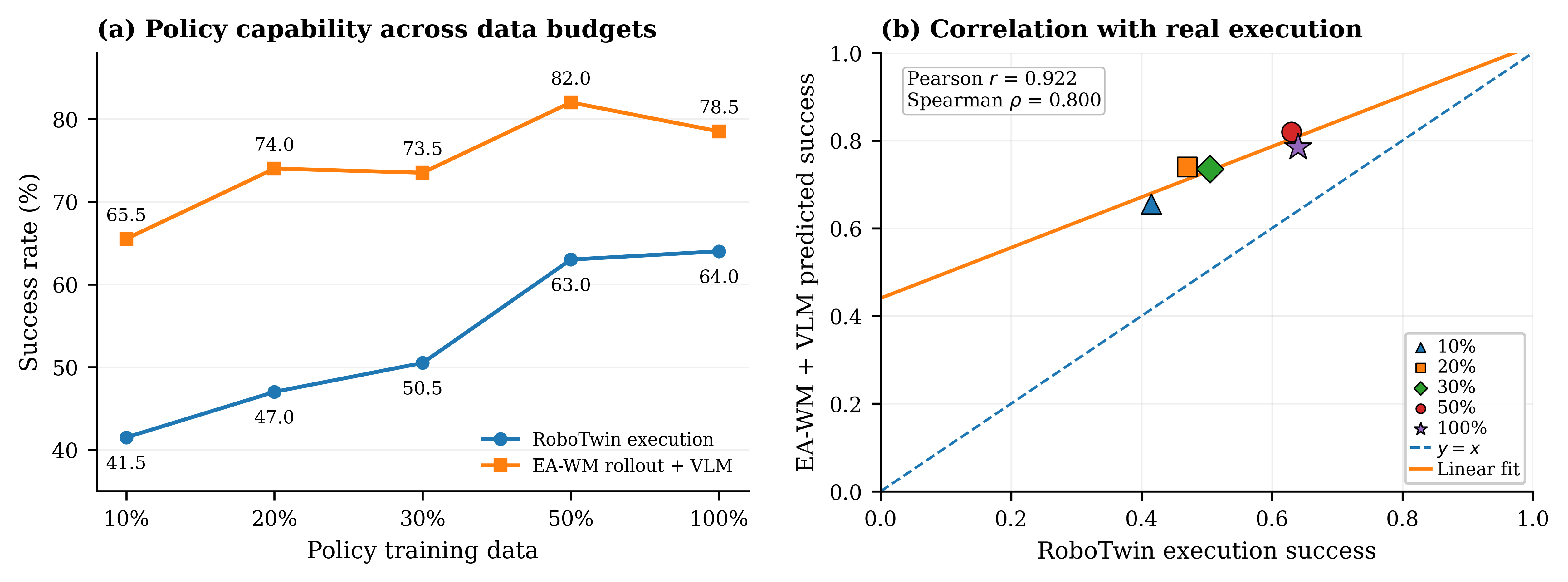}
    \caption{
    Policy evaluation with EA-WM.
    \textbf{(a)} RoboTwin execution success and EA-WM rollout-based evaluation for policies trained with different data budgets.
    \textbf{(b)} Correlation between real execution success and EA-WM rollout evaluation across the five policies. EA-WM reaches Pearson $r=0.922$ and Spearman $\rho=0.800$.
    }
    \label{fig:policy_evaluation}
\end{figure}

The rollout evaluator is somewhat optimistic in absolute success rate, but tracks policy capability consistently as the training-data budget changes. Its Pearson correlation is close to the strongest result reported by WorldArena (CtrlWorld, $r=0.986$) and substantially higher than Cosmos-Predict 2.5 ($r=0.483$). These results indicate that EA-WM rollouts provide a useful proxy for comparing policies with different capability levels.

\textbf{Action Planning.}
Finally, we use the temporal decoder to convert generated KVAF trajectories into executable actions and evaluate them directly in RoboTwin under the WorldArena action-planning protocol.

\begin{table}[H]
\caption{
Action-planning success on the two WorldArena tasks. EA-WM achieves the strongest average success among the compared world-model planners.
}
\label{tab:action_planning}
\centering
\small
\setlength{\tabcolsep}{8pt}
\begin{tabular}{lccc}
\toprule
Planner & \texttt{adjust\_bottle} & \texttt{click\_bell} & Average \\
\midrule
$\pi_{0.5}$ policy & 77\% & 66\% & 71.5\% \\
WoW & 20\% & 21\% & 20.5\% \\
TesserAct & 1\% & 35\% & 18.0\% \\
EA-WM KVAF planning & 35\% & \textbf{97\%} & \textbf{66.0\%} \\
\bottomrule
\end{tabular}
\end{table}

EA-WM reaches 66.0\% average planning success, substantially outperforming the other world-model planners in this comparison and approaching the direct $\pi_{0.5}$ policy reference. Performance is particularly strong on \texttt{click\_bell} (97\%), while the lower success on \texttt{adjust\_bottle} (35\%) indicates that precise multi-stage manipulation remains challenging. This gap is consistent with the fact that planning depends on both the accuracy of generated KVAF trajectories and the subsequent KVAF-to-action decoding process.

\subsection{Preliminary Real-World Evaluation on RH20T}
\label{sec:rh20t_realworld}

We further conduct a preliminary offline evaluation on RH20T to examine whether EA-WM remains effective beyond the simulated RoboTwin setting. We use data collected with a UR5 robot and WSG-50 gripper from a fixed global camera, which introduces both real-world appearance and a different robot embodiment from our main experiments. We construct 2,000 short clips for post-training and evaluate on 10 held-out episodes covering 10 tasks. Table~\ref{tab:rh20t_results} reports the six robot-centric metrics used by P3CScore together with the overall aggregate score.

\begin{table}[H]
\caption{
Preliminary real-world evaluation on RH20T. EA-WM achieves the best overall P3CScore, with the largest gains concentrated on interaction, trajectory, and depth modeling.
}
\label{tab:rh20t_results}
\centering
\small
\setlength{\tabcolsep}{3.8pt}
\begin{tabular}{lcccccccc}
\toprule
& \multicolumn{2}{c}{Physics Adherence}
& \multicolumn{2}{c}{3D Accuracy}
& \multicolumn{2}{c}{Controllability}
& & \\
\cmidrule(lr){2-3}
\cmidrule(lr){4-5}
\cmidrule(lr){6-7}
Model
& Inter.
& Traj.
& Depth
& Persp.
& Instr.
& Sem.
& P3CScore
& EWMScore \\
\midrule
CtrlWorld (w/o PT)
& 0.400
& 0.279
& 0.802
& 0.600
& 0.276
& 0.831
& 53.12
& 47.97 \\

Wan2.2 (w/o PT)
& 0.362
& 0.308
& 0.801
& 0.738
& 0.244
& 0.840
& 54.88
& 51.94 \\

CtrlWorld (PT)
& 0.468
& 0.507
& 0.889
& 0.798
& 0.342
& 0.867
& 64.51
& 54.41 \\

Wan2.2 (PT)
& 0.508
& 0.563
& 0.849
& \textbf{0.858}
& \textbf{0.378}
& \textbf{0.877}
& 67.22
& 56.51 \\

EA-WM (PT)
& \textbf{0.542}
& \textbf{0.608}
& \textbf{0.927}
& 0.808
& 0.374
& 0.866
& \textbf{68.75}
& \textbf{57.59} \\
\bottomrule
\end{tabular}

\vspace{2pt}
\end{table}

EA-WM achieves the best P3CScore of 68.75 and the highest overall EWMScore of 57.59. Compared with the strongest post-trained baseline, Wan2.2, the improvement is most pronounced on the metrics directly targeted by KVAF: Interaction Quality increases from 0.508 to 0.542, Trajectory Accuracy from 0.563 to 0.608, and Depth Accuracy from 0.849 to 0.927. EA-WM does not improve every individual metric, but its gains are concentrated on robot motion and 3D geometry, consistent with the trends observed in the RoboTwin experiments.

\begin{figure}[H]
    \centering
    \includegraphics[width=\linewidth]{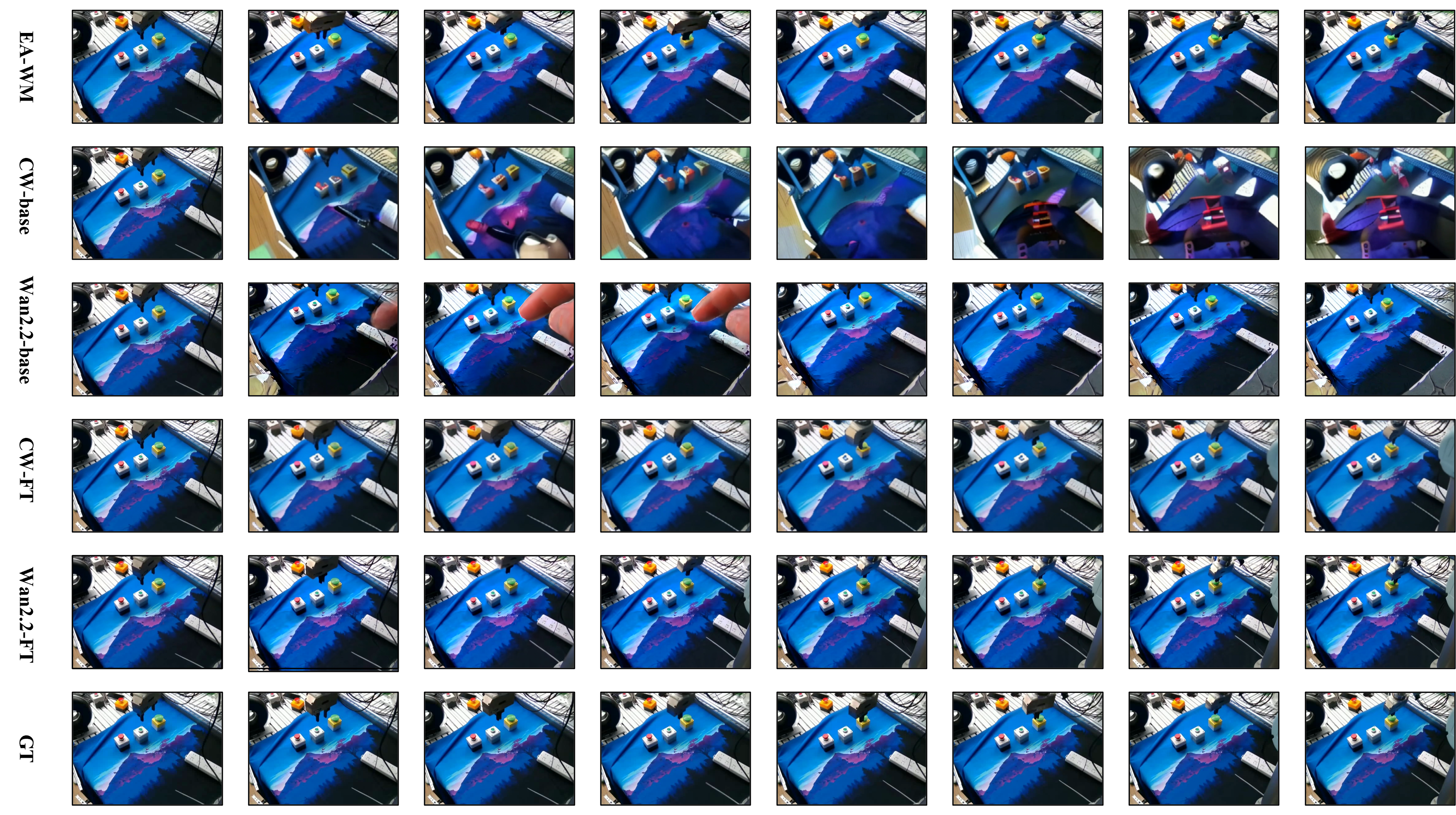}
    \caption{
    Qualitative rollouts on held-out RH20T episodes. The examples illustrate that EA-WM can preserve the overall real-world scene layout while generating plausible robot motion and interaction progress across different tasks. Occasional appearance and geometry drift remain in some rollouts, indicating that real-world generation is still imperfect.
    }
    \label{fig:rh20t_qualitative}
\end{figure}

The qualitative examples in Fig.~\ref{fig:rh20t_qualitative} show a similar trend. EA-WM can preserve the overall scene layout while generating plausible robot motion and interaction progress across held-out real-world tasks, although occasional appearance and geometry drift remain. Since this evaluation is limited to one robot embodiment, one fixed camera, 2,000 post-training clips, and 10 held-out episodes, we view these results as preliminary evidence of real-world transfer rather than a comprehensive evaluation of robustness or embodiment generalization.

\clearpage

\subsection{Counterfactual Condition Following}
\label{sec:counterfactual_following}

We further examine whether EA-WM generates futures that respond to the supplied conditions rather than relying primarily on the initial observation. We construct a counterfactual test in which the initial frame is kept fixed, while both the text instruction and KVAF trajectory are replaced by those from a different task. Specifically, we retain the initial observation from \texttt{beat\_block\_hammer} and replace its original condition with the instruction and KVAF trajectory from \texttt{adjust\_bottle}.

\begin{figure}[H]
    \centering
    \includegraphics[width=\linewidth]{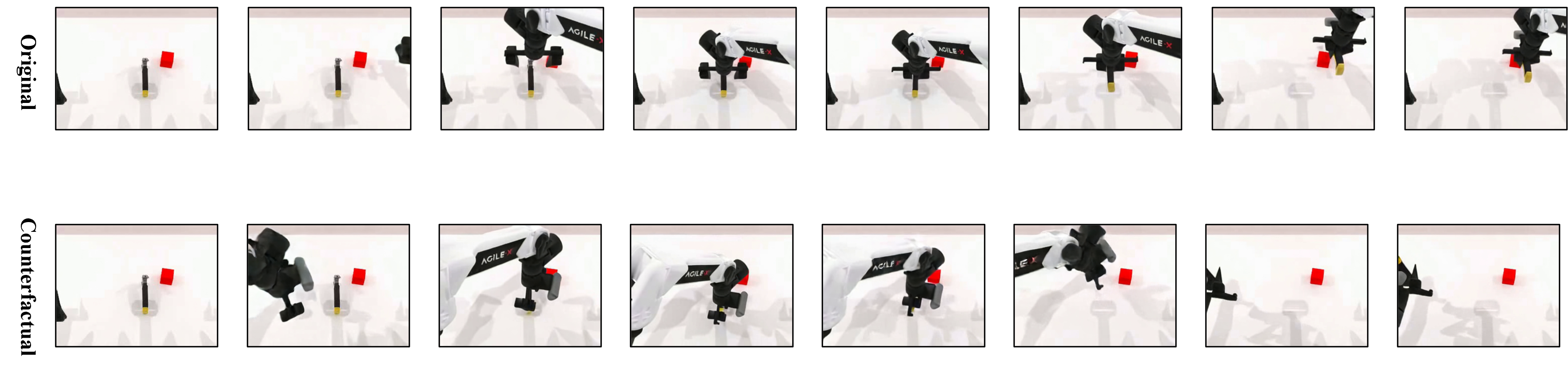}
    \caption{
    Counterfactual condition-following results. Both rollouts start from the same initial observation. The top row uses the original condition (\texttt{beat\_block\_hammer}), while the bottom row uses the counterfactual condition (\texttt{adjust\_bottle}). Changing the text and KVAF conditions leads to clearly different robot trajectories while preserving the initial scene context.
    }
    \label{fig:counterfactual_following}
\end{figure}

As shown in Fig.~\ref{fig:counterfactual_following}, the two rollouts share the same initial scene but quickly diverge in their subsequent robot motion. Under the original condition, the arm follows a trajectory concentrated around the original manipulation region, whereas the counterfactual condition induces a distinct approach and motion pattern. Importantly, the surrounding scene and object layout remain largely consistent while the robot trajectory changes, consistent with the model preserving scene context while responding to condition-dependent motion cues. This result provides qualitative evidence that EA-WM does not simply reproduce a canonical continuation associated with the first frame, but uses the supplied text and KVAF trajectory to modulate future robot motion and generate condition-dependent rollouts.

\end{document}